\documentclass[preprint,12pt]{elsarticle}

\usepackage{subcaption}
\usepackage{lineno,hyperref}
\usepackage{natbib}
\usepackage{geometry}
\usepackage{fleqn}
\usepackage{graphicx}
\usepackage{newtxtext,newtxmath}
\usepackage{hyperref}
\usepackage{booktabs}
\usepackage{bm}
\usepackage{amsmath}
\usepackage{enumitem}
\usepackage{threeparttable}
\usepackage{multirow}
\usepackage{esvect}
\usepackage{siunitx}
\usepackage{soul,xcolor}
\modulolinenumbers[5]

\graphicspath{{./figures/}}
\journal{Elsevier}

\usepackage{caption}
\usepackage[table]{xcolor}
\usepackage{siunitx}
\usepackage{etoolbox}
\appto\TPTnoteSettings{\footnotesize}

\begin{document}

\begin{frontmatter}

\title{\large{A Three-Stage Bayesian Transfer Learning Framework to Improve Predictions in Data-Scarce Domains}\tnoteref{t1}}

\tnotetext[t1]{Notice: This manuscript has been authored by UT-Battelle LLC under contract DE-AC05-00OR22725 with the US Department of Energy (DOE). 
The US government retains and the publisher, by accepting the article for publication, acknowledges that the US government retains a nonexclusive, paid-up, irrevocable, worldwide license to publish or reproduce the published form of this manuscript, or allow others to do so, for US government purposes. 
DOE will provide public access to these results of federally sponsored research in accordance with the DOE Public Access Plan (
\url{https://www.energy.gov/doe-public-access-plan})}

\author[NCSU]{Aidan Furlong\corref{mycorrespondingauthor}}
\cortext[mycorrespondingauthor]{Corresponding author}
\ead{ajfurlon@ncsu.edu}

\author[ORNL]{Robert K. Salko}

\author[UTK]{Xingang Zhao}

\author[NCSU]{Xu Wu}

\address[NCSU]{Department of Nuclear Engineering, North Carolina State University    \\ 
	Burlington Engineering Laboratories, 2500 Stinson Drive, Raleigh, NC 27695 \\}

\address[ORNL]{Nuclear Energy and Fuel Cycle Division, Oak Ridge National Laboratory \\ 1 Bethel Valley Road, Oak Ridge, TN 37830}

\address[UTK]{Department of Nuclear Engineering, University of Tennessee, Knoxville \\ Zeanah Engineering Complex, 863 Neyland Drive, Knoxville, TN 37916}

\begin{abstract}
The use of ML in nuclear engineering has grown steadily to support a broad array of applications. Among these methods, deep neural networks have been widely adopted due to their performance and accessibility, but they require large, high-quality datasets. Experimental data are often sparse, noisy, or insufficient to build resilient data-driven models. Transfer learning, which leverages relevant data-abundant source domains to assist learning in data-scarce target domains, has shown efficacy. Parameter transfer, where pretrained weights are reused, is common but degrades under large domain shifts. Domain-adversarial neural networks (DANNs) help address this issue by learning domain-invariant representations, thereby improving transfer under greater domain shifts in a semi-supervised setting. However, DANNs can be unstable during training and lack a native means for uncertainty quantification.

This study introduces a fully-supervised three-stage framework, the staged Bayesian domain-adversarial neural network (staged B-DANN), that combines parameter transfer and shared latent space adaptation. In Stage 1, a deterministic feature extractor is trained on the source domain. This feature extractor is then adversarially refined using a DANN in Stage 2. In Stage 3, a Bayesian neural network is built on the adapted feature extractor for fine-tuning on the target domain to handle conditional shifts and yield calibrated uncertainty estimates. This staged B-DANN approach was first validated on a synthetic benchmark, where it was shown to significantly outperform standard transfer techniques. It was then applied to the task of predicting critical heat flux in rectangular channels, leveraging data from tube experiments as the source domain. The results of this study show that the staged B-DANN method can improve predictive accuracy and generalization, potentially assisting other domains in nuclear engineering.
\end{abstract}

\begin{keyword}
transfer learning, domain adaptation, hybrid modeling, critical heat flux, uncertainty quantification
\end{keyword}

\end{frontmatter}

\section{Introduction}

The rise of advanced ML over the past decade has given way to a diverse set of mathematical frameworks designed to make predictions in a given task by using related information previously seen. Nearly all these techniques make the simple assumption that the data used to train the model are from the same feature space and distribution as the prediction domain---an assumption that in many cases is flawed~\cite{niu2021decade}. Additionally, there are many instances where there are insufficient amounts of clean data to train on, making many tasks and implementations of ML infeasible. To counter these problems, a family of methods known as \textit{transfer learning} (TL) was developed~\cite{bozinovski1976influence}\cite{pan2009survey}. The central idea behind TL is that knowledge obtained previously in one model can benefit another model in a similar task or domain. TL has the potential to offer several practical benefits to a target model: an increased rate of model training convergence, reduced training data requirements, and an overall increase in performance on the target task.

The simplest transfer strategy, \textit{parameter transfer}~\cite{zhuang2020comprehensive}, assumes that the weights of a model pretrained on a \textit{source domain} encode generally useful representations that can be readily applied to a more limited \textit{target domain}. Freezing or fine-tuning the pretrained layers can produce rapid convergence during training but may propagate misleading input--output relationships if larger distributional differences exist between domains. This can lead to convergence in suboptimal minima and even result in negative transfer, where performance is below that of a model trained from scratch on the target domain. These failures can be particularly evident in deep networks because later pretrained layers capturing task-specific correlations may actively hinder the target domain training process~\cite{zhang2022survey}.

To mitigate some concerns regarding covariate shift, domain-adversarial neural networks (DANNs) have been developed. DANNs enforce feature invariance via adversarial alignment~\cite{ganin2015unsupervised}. DANNs were originally developed for computer vision applications in unsupervised domain adaptation but have subsequently been extended~\cite{farahani2021domain} to regression problems and applied in engineering applications. Although DANNs are effective, their predictions remain deterministic and exhibit instabilities during training. The adversarial training dynamic common to DANNs often leads to oscillatory or collapsed representations unless hyperparameters are extensively tuned and the joint optimization process is well-behaved~\cite{acuna2022domain}. The deterministic nature of DANNs also limits their application in sensitive or safety-critical engineering contexts because no predictive uncertainties are produced. Without reliable confidence estimates, it is difficult to establish model trustworthiness or create selective model implementations~\cite{wu2025uncertainty}.

In this work, a TL strategy was developed that combines the benefits of parameter transfer, adversarial domain alignment, and Bayesian neural network (BNN) architectures into a modular three-stage framework. Stage 1 of this approach involves training a deterministic feature extractor on the source domain, establishing initial representations and encouraging downstream stability. In Stage 2, a DANN architecture aligns the source and target distributions via adversarial training, isolating domain-invariant features. In Stage 3, the deterministic components are replaced with a BNN for fine-tuning on target data, adapting the model to conditional shift and providing uncertainty awareness~\cite{gawlikowski2023survey}. This staged Bayesian domain-adversarial neural network (staged B-DANN) framework was first validated on a synthetic benchmark designed to simulate both covariate and conditional shift in a controlled environment. It was then applied on a novel thermal hydraulics application predicting critical heat flux (CHF) in rectangular channels, leveraging tube-based datasets for cross-geometry transfer.

By structuring the staged B-DANN to integrate representation alignment and Bayesian inference in a cohesive TL strategy, the framework offers a high-performance, uncertainty-aware solution for small-data scenarios. The method was experimentally validated on two representative case studies, demonstrating its performance gains over other training strategies and its applicability to real-world engineering problems.

The remainder of this paper is structured as follows. Section~\ref{sec:background} reviews relevant TL concepts and establishes problem background. Section~\ref{sec:methods} details the staged B-DANN framework and associated optimization techniques. Section~\ref{sec:results} presents the case studies and performance comparisons between the staged B-DANN and common methods. Finally, Section~\ref{sec:conclusions} concludes with major findings and future directions.

\section{Background}
\label{sec:background}

Like many recently popularized ML techniques, TL is not a new concept. The original proposal of such an idea by S.~Bozinoski~\cite{bozinovski1976influence} in 1976 considered the direct transfer of information between neural networks to bolster the performance in a similar-but-different domain. This idea was then further developed with a more formal theoretical foundation by Pratt and Thrun in 1997 \cite{pratt1997special}. Today, TL is an active component in many tasks, such as image recognition, medical imaging, gameplay, and missile target acquisition \cite{pan2009survey}.

\subsection{Fundamentals of Transfer Learning}
\label{subsec:transfer_learning}

The following is a general, formalized definition of the concept of TL from Weiss et~al.~\cite{weiss2016survey}: 

\begin{quote}
    ``Given a source domain $\mathcal{D}_S$ with a corresponding source task $\mathcal{T}_S$ and a target domain $\mathcal{D}_T$ with a corresponding target task $\mathcal{T}_T$, transfer learning is the process of improving the target predictive function $f_T(\cdot)$ by using the related information from $\mathcal{D}_S$ and $\mathcal{T}_S$, where $\mathcal{D}_S \neq \mathcal{D}_T$ or $\mathcal{T}_S \neq \mathcal{T}_T$.''
\end{quote}

A domain is defined by two parts: $\mathcal{D} = \{\mathcal{X}, P(X)\}$, with $X = \{x_1,x_2,...,x_n\} \in \mathcal{X}$ representing a particular learning sample within a feature space $\mathcal{X}$. Here, $P(X)$ denotes the marginal probability distribution of observations $X$. For a given $\mathcal{D}$, a task can be defined by two parts: $\mathcal{T} = \{\mathcal{Y}, f(\cdot)\} = \{\mathcal{Y}, P(Y | X)\}$, where $Y = \{y_1,y_2,...,y_n\} \in \mathcal{Y}$ is the label space, and $P(Y | X)$ is the conditional probability distribution of $Y$ given $X$ or how likely certain labels are for each input. Having different source/target domains $\mathcal{D}_S \neq \mathcal{D}_T$ means that $\mathcal{X}_S \neq \mathcal{X}_T$ and/or $P_S(X) \neq P_T(X)$. Specifically, the definition of TL implies that one or more of the following is true:

\begin{itemize}
    \item The feature spaces are different ($\mathcal{X}_S \neq \mathcal{X}_T$).
    \item The label spaces are different ($\mathcal{Y}_S \neq \mathcal{Y}_T$).
    \item The marginal probability distributions are different ($P_S(X) \neq P_T(X)$).
    \item The conditional probability distributions are different ($P_S(Y | X) \neq P_T(Y | X)$).
\end{itemize}

When only the probability distributions differ ($P_S(X) \neq P_T(X)$ or ($P_S(Y | X) \neq P_T(Y | X)$), the scenario is referred to as \textit{domain adaptation}. The term \textit{homogeneous} TL refers to cases where the input feature spaces are the same between source and target domains ($\mathcal{X}_S = \mathcal{X}_T$) and can also have differences in distributions. As a practical example, a model trained to recognize objects in a photograph (source domain) could, via TL, be adapted for use in the analysis of X-ray images with similar feature representations (target domain). Conversely, \textit{heterogeneous} TL denotes cases where the feature spaces are different ($\mathcal{X}_S \neq \mathcal{X}_T$) and the probability distributions are different, such as transferring from image data to text data.

To classify the various approaches within TL, categorization by problem (\textit{label setting}) or categorization by solution (\textit{transfer setting}) must be considered. In the label setting, solutions can be further categorized~\cite{pan2009survey}. \textit{Inductive} approaches deal with dissimilar source and target tasks but similar domains. In \textit{transductive} strategies, the tasks are the same, but the domains are different. Finally, \textit{unsupervised} TL problems have both dissimilar tasks and domains, and without abundant labels to use during training.

In the transfer setting perspective, TL approaches can be differentiated into four major areas, which can be members of multiple label setting categories~\cite{tan2018survey}:
\begin{itemize}
    \item \textit{Instance-transfer}: Re-weight a portion of labeled data in the source domain for use in the target domain.
    \item \textit{Mapping-transfer}: Find a feature representation that reduces difference between the source and target domains as well as the error of classification/regression.
    \item \textit{Model-transfer}: Discover shared model parameters or priors between the source domain and target domain models that can benefit from TL (\textit{network}-based).
    \item \textit{Relational-knowledge-transfer}: Build mapping of relational knowledge between the source and target domains. Both domains are relational, and the i.i.d.~assumption is relaxed.
\end{itemize}

\subsection{Parameter Transfer}
\label{subsec:parameter_transfer}

The most common (and most easily implemented) form of TL, parameter transfer~\cite{zhuang2020comprehensive}, is where the parameters of a well-trained source model are used as the initialization or foundation for the target model. A member of the model-transfer method group, this approach relies on the assumption that there are similarities in underlying structures between source and target models. In parameter transfer, the source model is typically composed of modular components such as a feature extractor and regression head (for regression tasks), both of which are trained on the data from a different-but-related source domain. Once these are finalized, the model can be transferred to the target domain via one of the following strategies:

\begin{itemize}
\item \textit{Frozen reuse}: The source-trained components are reused without modification with new target-specific layers trained on top.
\item \textit{Partial fine-tuning}: Selected layers are unfrozen and adapted to the target domain, whereas others remain fixed.
\item \textit{Full fine-tuning}: All transferred layers are initialized from the source model and re-trained on the target data.
\end{itemize}

Selection of one of the above strategies is often informed empirically and methodologically. One such method is performed by continually unfreezing layers from the output of the feature extractor until validation loss fails to improve or worsens. This process is rooted in the fundamental idea of layerwise generality, where early hidden layers tend to contain coarse, domain-agnostic features, whereas deeper layers contain finer, domain-specific information~\cite{yosinski2014transferable}. Ideally, the transferred parameters will encode representations common to both domains. 

Even when unfreezing the entire model's parameters and allowing the complete re-training of all layers, information can be transferred via informed parameter initialization. Ordinarily, the weights and biases of a model are initialized with random values; this determines the starting location in the loss landscape. This randomness has limited effect in large-data regimes and highly generalizable models, but it becomes increasingly consequential when data are scarce or the model is overparameterized. Due to this sensitivity, a more informed initialization should provide an increase in convergence rate and model generalization~\cite{erhan2010does}. A straightforward way of achieving this is full fine-tuning, where the source model's (pretrained) components are used as initializations, and the entire target model is re-trained with target domain data.

Although it is a common method, parameter transfer does have known limitations. It typically assumes that the conditional distribution remains consistent between source and target domains ($P_S(Y | X) = P_T(Y | X)$). When this assumption is invalid, such as in cases of conditional shift, the transferred parameters may possess misleading or incorrect input--output relationships, especially in deeper layers. Additionally, parameter transfer does not enforce alignment between the marginal feature distributions ($P(X)$), leading to potentially poor performance when there is a high covariate shift. In both of these cases, fine-tuning may converge to suboptimal solutions, degrading performance potentially to the point of \textit{negative transfer}; this is observed when the performance of the TL-aided model decreases below that of a target model trained from scratch.

\subsection{Domain-Adversarial Neural Networks}
\label{subsec:danns}

To mitigate these potential issues, several alternative TL techniques have been proposed, such as feature-based domain adaptation methods. These approaches explicitly encourage the learning of domain-invariant feature representations, aligning the source and target feature distributions within the latent space and improving generalizations in covariate shift. One such method is the DANN~\cite{ganin2016domain}, originally designed for classification tasks. This method considers a labeled source dataset and an unlabeled target dataset, meaning that it is defined as a semi-supervised method.

The canonical DANN consists of three primary components: a feature extractor, label predictor, and domain classifier. The feature extractor maps inputs to a shared latent space. From this representation, the label predictor is trained to predict the output (e.g., class membership) using only the labeled source data. The predictor's loss is computed from the prediction error and backpropagated to update both the label predictor and the feature extractor. In parallel, the domain classifier attempts to discriminate between source and target domain samples. This classifier's loss is also computed and backpropagated through the classifier's parameters but then passes through a gradient reversal layer (GRL) situated between the domain classifier and the feature extractor. This layer negates and scales the gradient signal, creating a true adversarial and competitive relationship between domain classifier and label predictor. Ideally, this promotes the feature extractor to produce domain-invariant features that ``confuse'' the domain classifier. This design allows the model to learn features that are both discriminative for the source task and invariant to domain-specific information, thereby improving generalization to the unlabeled target domain.

The applications of the DANN approach have predominantly been in computer vision, but investigation has also been active in other fields~\cite{dai2023smart}\cite{wang2025transfer}. A modification of this framework, DANN-R~\cite{farahani2021domain}, extends this concept to regression tasks and validates its use on physical data from turbine instruments. The primary change to the original DANN concept is to replace the label predictor stack with a regression stack and change the associated loss function to mean squared error (MSE) instead of the two-class binary cross entropy objective. Farahani et al.~\cite{farahani2021domain} compared the DANN-R method to counterpart models without TL on physical data and showed that their TL approach can significantly outperform ordinary deep neural network (DNN) models. 

Another method, known as adversarial discriminative domain adaptation (ADDA)~\cite{tzeng2017adversarial}, uses a decoupled approach for handling encoders. Instead of having a shared feature extractor for both the source and target domains, it instead separates them and uses a two-step training procedure: pretraining and alignment. The source domain encoder is first pretrained using only the labeled source domain data and is subsequently frozen. The frozen extractor is then used in parallel with a trainable target domain encoder in the adversarial stage such that a discriminator cannot correctly predict the domain label. ADDA outperformed three other domain adaptation methods, including DANNs, on a set of image-based validation tests.

\section{Methods}
\label{sec:methods}

This section presents the proposed staged B-DANN approach used to transfer knowledge of a source domain to a target domain using a modular three-stage architecture. The methodology integrates supervised regression, adversarial domain adaptation, and Bayesian fine-tuning to address both covariate and conditional shift.

\subsection{Staged B-DANN}
\label{subsec:staged_b-dann}

\begin{figure}[htb!]
    \centering
    \includegraphics[width=\linewidth]{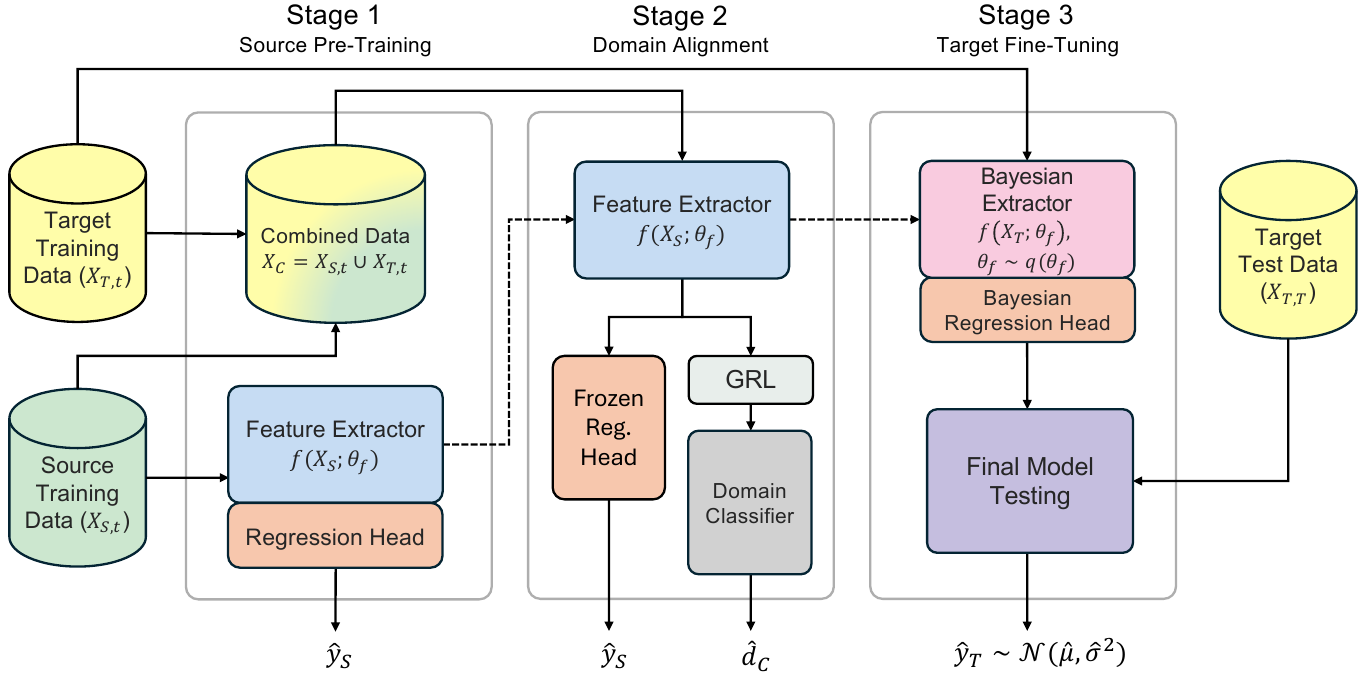}
    \caption{Training workflow for the three-stage TL approach. Note that the stage two $\hat{y}_S$ output is only used for diagnostic purposes and does not affect training gradients.}
    \label{fig:bdann_workflow}
\end{figure}

The design of the staged B-DANN approach lies in a three-stage process, as shown in Figure \ref{fig:bdann_workflow}. The staged B-DANN approach can be described as being both a mapping-transfer solution and a model-transfer solution according to the taxonomy of TL methods defined in Section \ref{subsec:transfer_learning}; the strategy finds a shared feature representation and then fine-tunes it in the target domain by sharing the domain-invariant model parameters.

The core process of the staged B-DANN is the continual refinement of the shared feature extractor over the three stages. The first stage is a simple DNN trained on source domain training data ($X_{S,t}$), which are preprocessed using standard practices. In this diagram, the first subscript of the data variables indicates whether they are from the source ($S,$) or target ($t,$) dataset, and the second subscript represents whether they are a training ($,t$) or testing ($,T$) partition. Post training, a test partition of the source data is used to provide an evaluation of the model's ability to fulfill the source task. The feature extractor is then transitioned to Stage 2, which deals with the domain alignment. 

Stage 2 performs alignment using both the source and target data, which are combined and assigned binary domain labels (0 for source or 1 for target) to denote class membership, making this process supervised. This stage's architecture has two branches: (1) the adversarial domain classifier paired with the GRL to encourage domain-invariant features and (2) the frozen regression head from the pretrained Stage 1 model. The adversarial domain classifier outputs a predicted class label ($\hat{d}_C$) for a given input vector, and the frozen regression head produces source domain predictions ($\hat{y}_S$). The regression head is not trainable and does not contribute to the total loss function or gradient updates of this stage. This design decision mitigates the instability concerns noted in Section \ref{subsec:danns} and does not negatively affect downstream processes because the feature extractor is re-trained in Stage 3. The inclusion of the regression head is still valuable for monitoring source predictions; it ensures that source domain performance does not degrade due to catastrophic forgetting.

Once the domain classifier's accuracy has converged (ideally around 50\% to indicate maximal confusion), the deterministic feature extractor is transferred to Stage 3. It is then used to initialize the parameters of a variational BNN, replacing the default zero-mean Gaussian priors with the informative point-value weights of the deterministic feature extractor~\cite{krishnan2020specifying}. The frozen regression head from Stage 1 is also converted to a Bayesian variant and made trainable. 

The constructed BNN is then fully trained via variational inference using only target domain data, forming posterior distributions over the weights. During inference, predictions for both means and standard deviations are made via Monte Carlo sampling from these distributions. The resulting means ($\hat{\mu}$) of these sampled means provide final point-value estimates ($\hat{y}_T$), and the standard deviations ($\hat{\sigma}$) of these sampled means provide the associated epistemic uncertainties. The standard deviations directly predicted by the BNN represent the aleatoric uncertainty, which forms the total uncertainty when combined with the epistemic uncertainty.

\subsubsection{Stage 1: Pretraining}
\label{subsubsec:stage1}

The first training process of the staged B-DANN pipeline involves the supervised training of a deterministic DNN using only the source domain data. This stage serves to initialize the feature extractor with a representation that captures the input--output relationships of the source task and provides stability for subsequent adversarial training.
Empirical evidence indicates that this kind of feature extractor pretraining can improve both predictive performance and optimization stability~\cite{tzeng2017adversarial}. A pretrained model offers more informed weight initializations than default random values, helping to mitigate instability challenges known in adversarial architectures.

Any standard supervised loss function may be used for the Stage 1 DNN. In this work, the MSE loss was selected due to its simplicity and common usage in regression problems. Formally, this loss is defined in Eq.~(\ref{eqn:stage1_loss}), where $N_s$ denotes the number of source samples, $y_i^{(s)}$ denotes the true value for source sample \textit{i}, and $\hat{y}_i^{(s)}$ denotes the predicted value for source sample \textit{i}.

\begin{equation} \label{eqn:stage1_loss}
\mathcal{L}_{\text{Stage 1}} = \frac{1}{N_s} \sum_{i=1}^{N_s} \left( y_i^{(s)} - \hat{y}_i^{(s)} \right)^2
\end{equation}

After training, the feature extractor is passed to Stage 2 for domain alignment. A small source testing partition may be used to verify basic task competency before domain adaptation.

\subsubsection{Stage 2: Domain Alignment}
\label{subsubsec:stage2}

Once the feature extractor and regression head are transferred from Stage 1, the second stage addresses domain alignment by encouraging the extractor to learn domain-invariant representations. This is done through adversarial training against the domain classifier, attempting to distinguish between the source and target domains. Both the feature extractor and domain classifier are trainable, with the GRL in between, updating the feature extractor's weights opposite of the gradients computed via the domain classifier's loss function. The effective goal in doing this is to confuse the classifier, increasing its loss and converging its accuracy to 50\%. The regression head from Stage 1 is frozen and not trainable in this stage, leaving the total loss equivalent to the binary cross entropy (BCE) of the domain classifier, as shown in Eq.~(\ref{eqn:stage2_loss}). The subscript ``DC'' denotes the domain classifier:

\begin{equation} \label{eqn:stage2_loss}
\mathcal{L}_{\text{Stage 2,DC}} = \left( -\frac{1}{N_s + N_t} \sum_{i=1}^{N_s + N_t} \left[ d_i \log(\hat{d}_i) + (1 - d_i) \log(1 - \hat{d}_i) \right] \right)
\end{equation}

Here, $d_i \in \{0,1\}$ is the true domain label of sample \textit{i}, where 0 is the source domain, and 1 is the target domain. The variable $\hat{d}_i \in (0, 1)$ is the predicted domain probability. The variables $N_s$ and $N_t$ represent the number of samples in the source and target classes, respectively. Defining the GRL is simple, which merely flips the sign of the DC loss and scales it to create the Stage 2 feature extractor loss (``FE''):

\begin{equation}
    \mathcal{L}_{\text{Stage 2,FE}} = -\lambda(e) \cdot \mathcal{L}_{\text{Stage 2,DC}}
\end{equation}

The variable $\lambda(e)$ is the domain loss weighting at training epoch \textit{e}. This value can be held static but performs best when scheduled with respect to the training progress, which is computed from epoch number and other prescribed coefficients further discussed below. One limitation of Stage 2 is a problem-dependent training instability and sensitivity to choice of hyperparameters. This sensitivity is a known aspect of adversarial architectures~\cite{acuna2022domain}. The adversarial characteristics of conventional DANNs, involving the combination of gradient descent and joint optimization to find a saddle point, are inherently unstable and contribute greatly to the overall instability of the method~\cite{mazumdar2020gradient}. Compared with traditional DANNs, the staged B-DANN's Stage 2 regression head is frozen and does not contribute to the total loss or gradients. Therefore, training does not perform a joint optimization, so the above concerns are mitigated by design. For further mitigation, several strategies exist: lambda scheduling~\cite{ganin2015unsupervised}, domain classifier early stopping, and batch balancing.

The first of these mitigation strategies is the lambda scheduling method, which continually increases the lambda parameter over the training period to a prescribed maximum value. This sequentially increases the overall Stage 2 loss and subsequently the backpropagated gradients, leading to more significant weight updates. This is done to gradually introduce the adversarial influence over time; if the initial influence is too large, instability may arise even if the initializations are well-formed from Stage 1's pretraining. The scheduler chosen follows a modification of the form proposed by Ganin et al.~\cite{ganin2015unsupervised}:

\begin{equation} \label{eqn:lambda_scheduling}
\lambda(e) =
\begin{cases}
0, & e < e_{\mathrm{warm}}, \\
\lambda_{\max} \cdot \,
\max\!\left(
\lambda_{\min},\,
\dfrac{2}{1 + \exp[-k\,p(e)]} - 1
\right), & e \ge e_{\mathrm{warm}},
\end{cases}
\end{equation}

where the normalized progress variable $p(e)$ is defined as
\begin{equation}
p(e) = \frac{e - e_{\mathrm{warm}}}{\max(1,\,E_{\mathrm{tot}} - e_{\mathrm{warm}})}.
\end{equation}

For epochs $e < e_{\mathrm{warm}}$, the adversarial term is disabled to stabilize the feature extractor and regressor. After this warmup period, $\lambda(e)$ rises smoothly toward $\lambda_{\max}$ following a logistic ramp with a steepness control $k$, and optionally maintaining a nonzero floor $\lambda_{\min}$ to prevent vanishing gradients. Variable $E_{\mathrm{tot}}$ defines the maximum number of epochs prescribed at the beginning of training.

Implementing an early stopping protocol is a necessary step to reduce the potential for overfitting. Simple approaches to this typically involve the monitoring of a single variable such as classifier accuracy or validation loss to determine if the model is stable and converged. Given the nature of the Stage 2 architecture, use of BCE validation loss alone was inadequate, and experienced false stoppage simply due to intermittent instability. To guard against this, a composite monitoring function was developed that incorporates the area under the receiver operating characteristic curve (AUC). This function, $S_{\mathrm{val}}$, can be fully defined as

\begin{equation} \label{eqn:early_stopping}
S_{\mathrm{val}} =
\bigl|\mathrm{AUC}^{(\mathrm{val})} - 0.5\bigr|
+ \gamma \cdot \, \max\!\bigl(0,\,
\ln 2 - \mathcal{L}^{(\mathrm{val})}_{\mathrm{Stage2,DC}}\bigr),
\end{equation}

To formulate Eq.~(\ref{eqn:early_stopping}) for minimization, the $\mathrm{AUC}^{(\mathrm{val})}$ has 0.5 subtracted, since 0.5 indicates complete confusion of the DC. Similarly, $\ln 2$ is the loss of the BCE function corresponding to complete randomness, also indicating the classifier's confusion. The scaling term $\gamma$ is prescribed, which was set to 0.5 for this study.

Because one of the goals of this framework is to use a source dataset that is much larger than the target dataset, considerations must be made regarding batch composition for the domain classifier~\cite{sun2022source}. Large batch imbalances have the potential to bias the classifier toward the majority class, simply because there are more instances. Because Stage 2's domain labels are available, this process is relatively straightforward; the batch members are sampled from the source and target domains to enforce a one-to-one ratio, with replacement permitted for the smaller target dataset.

\subsubsection{Stage 3: Fine-Tuning}
\label{subsubsec:stage3}

In BNNs trained via variational inference, the weights are modeled as probability distributions instead of point estimates. In the case of a Gaussian variational posterior, the means and standard deviations are learned parameters. BNNs are designed similarly to DNNs in architecture; hidden layers control depth, and the number of neurons per layer controls width. With each neuron as a distribution, BNNs can implicitly capture uncertainties in both the data (aleatoric) and in the model itself (epistemic). 

When initialized, BNN weights are typically drawn from noninformative priors such as standard Gaussian distributions, which may place the model in a poor region of the loss landscape. This is of particular concern in low-data regimes. Improving these priors (and subsequently model performance) can be done by using the refined weights from the Stage 2 feature extractor. These point-value weights are taken and used to initialize the BNN's weight distributions as mean values, with the standard deviations set to a value of 0.1 to reflect a moderate-high confidence in these weights. By default, TensorFlow Probability initializes with a narrow standard deviation of 0.05, which may deteriorate performance in situations with large shift.

The BNN model is then fine-tuned using only the target domain data and trained with the variational objective. Stage 3 uses the evidence lower bound (ELBO) loss function in Eq.~(\ref{eqn:stage3_loss}). This loss function uses the negative log-likelihood of the observed target values under a predicted Gaussian output distribution, which is then regularized by a Kullback--Leibler (KL) divergence penalty between the prior ($p(\boldsymbol\theta)$) and the variational approximation to the posterior ($q(\boldsymbol\theta)$) over the model weights.

\begin{equation} \label{eqn:stage3_loss}
\mathcal{L}_{\text{Stage 3}} = \frac{1}{N_t} \sum_{i=1}^{N_t} \left[ \frac{1}{2} \log(2\pi \hat{\sigma}_i^2) + \frac{(y_i - \hat{y}_i)^2}{2 \hat{\sigma}_i^2} \right] + \frac{\beta(p)}{N_t} \, \mathcal{D}_{\mathrm{KL}} \left( q(\boldsymbol\theta) \, \| \, p(\boldsymbol\theta) \right)
\end{equation}

Here, $y_i$ is the true target value, $\hat{y}_i$ is the predicted mean, and $\hat{\sigma}^2_i$ is the predicted variance of the model's output for target instance $i$. Variable $\beta$ is a regularizer that weights the KL divergence term and effectively controls the amount of learning pressure during training~\cite{higgins2016early}. To improve convergence, an annealing strategy is used for $\beta$, which follows the same sigmoid form as the lambda scheduling from Eq.~(\ref{eqn:lambda_scheduling}):

\begin{equation}
\beta(p) = \beta_{\mathrm{max}} \cdot \frac{2}{1 + \exp\left(-10 \cdot p\right)} - 1
\end{equation}

Here, $\beta_{\mathrm{max}}$ is a prescribed value that is an important hyperparameter to tune for optimal performance. Once the BNN has completed training, the predictive distribution of a given input vector is approximated by drawing multiple stochastic forward passes. During each pass, two values are produced: the predicted mean and the variance of the output distribution. These predicted means are treated as samples from the underlying predictive distribution; the average of these means represents the final point estimate for each input vector, and the standard deviation of the means represents the associated epistemic uncertainty. The aleatoric uncertainty is separately quantified by averaging the variance outputs across the forward passes.

\subsection{Hyperparameter Optimization}
\label{subsec:hp_optimization}

Hyperparameter tuning and model optimization is an essential step in creating the best possible models for comparison. Even though training stability in the staged B-DANN method is greater than that of a standalone DANN, there is still a notable sensitivity to hyperparameters, particularly in Stage 2's $\lambda_{\mathrm{max}}$ and classifier architecture. To address this, automated hyperparameter optimization was performed in this study using the Optuna package~\cite{optuna_2019}. Optuna provides a structured framework for tuning and supports Bayesian optimization via surrogate modeling of the objective function.

For the staged B-DANN, a total of 80 configurations were trialed per model, optimizing for MSE evaluated on a validation partition. A 20-trial random search was conducted to improve initial exploration of the search space. With three stages, the dimensionality of the search space is high, so it was split into two smaller optimization subproblems. The first of these problems considered the architectural hyperparameters (e.g., hidden layers, widths, activation functions), which were tuned and fixed, and the second subspace included training-specific hyperparameters.

The search spaces for the DNN and BNN elements of the staged B-DANN were restricted to practical, commonly used ranges. The hyperparameter search space for the less common Stage 2 domain classifier is included in Table \ref{tab:staged_bdann_hp_tuning}. To reduce the search space dimensionality, the domain classifier's hidden layers were set using ReLU activations, with a sigmoid activation function for the output as appropriate for binary classification.

\begin{table}[htb!]
    \normalsize
    \captionsetup{justification=centering}
    \centering 
    \caption{Stage 2 domain classifier hyperparameter search space.}
    \label{tab:staged_bdann_hp_tuning}
    \begin{tabular}{ll}
        \toprule
        Hyperparameters & Ranges/values \\ 
        \midrule
        Domain classifier layers & \{1, 2, 3, 4\} \\ \midrule
        Domain classifier neurons & $[32, 256]$ \\ \midrule
        Dropout rate & $[0.0, 0.5]$ \\ \midrule
        Initial learning rate & $[10^{-5}, 10^{-4}]$ \\ \midrule
        Maximum $\lambda$ & $[0.1, 2.0]$ \\ \midrule
        Minimum $\lambda$ fraction & $[0.01,0.2]$ \\ \midrule
        Ramp \textit{k} & $[5,20]$ \\ \midrule
        Regularizer (L\textsubscript{2}) & $[10^{-7}, 10^{-3}]$ \\ \midrule
        Warmup epochs & $[0,15]$ \\
        \bottomrule
    \end{tabular}
\end{table}

\section{Case Studies}
\label{sec:results}

In this section, two studies are presented to evaluate the staged B-DANN method: a synthetically generated benchmark and a real-world nuclear engineering problem involving the prediction of CHF. Across both of these studies, three training strategies were implemented for analysis:

\begin{itemize}
    \item \textit{From-Scratch}: A single DNN trained exclusively on target domain data
    \item \textit{Direct Transfer}: A DNN pretrained on source domain data and then selectively fine-tuned on target data
    \item \textit{Staged B-DANN}: The proposed three-phase approach involving source pretraining, domain-adversarial alignment, and Bayesian fine-tuning, as described in Section~\ref{subsec:staged_b-dann}
\end{itemize}

The from-scratch and direct transfer baselines were chosen for interpretability and to bound performance achievable without a dedicated domain alignment stage. From-scratch training of a DNN using only target domain data provides a reference for how well the target task can be predicted without assistance from the source domain. Direct transfer, as noted in Section \ref{subsec:parameter_transfer}, is one of the most popular transfer learning methods and serves to test the usefulness of the inductive bias yielded from the source domain. It is known to be effective when the source and target conditional distributions are similar, but can see performance degradation with increasing conditional shift. These baselines allow for contributions made by the staged B-DANN to be readily identifiable: differences against from-scratch training indicate value over target-only learning, and improvements over direct transfer indicate that explicit alignment and staged adaptation provides additional value.

To provide an appropriate comparison, measures are necessary to isolate the advantages that the staged B-DANN provides. In each case study, models were made structurally identical with the exception of stage 2 of the B-DANN, which performs the domain alignment. The feature extractor and regression head architectures were held constant across the three strategies to eliminate performance differences simply due to their capacities. For both two TL-enabled approaches, the same source domain-trained base model was used for initialization. During training for each case study's experimental groups, all training, validation, and testing data was identical across methods with global determinism and fixed random seeds to ensure reproducibility and control for potential data handling variability.

All models were implemented using TensorFlow~\cite{tensorflow2015-whitepaper}. Training across all strategies used early stopping to mitigate overfitting concerns, with a maximum of 400 epochs and a patience of 20 epochs for the deterministic DNN and Bayesian variants. The Stage 2 domain classifier of the staged B-DANN was allowed a maximum of 100 epochs using the early stopping method identified in Eq.~(\ref{eqn:early_stopping}). An exponential learning rate decay was also used in the models, with a decay factor of 0.96 applied every 10 epochs.

\subsection{Case Study 1: Synthetic Benchmark}
\label{subsec:synthetic_benchmark}

The synthetic benchmark serves as a controlled environment for evaluating the three training strategies under known conditional shift, enabling a direct comparison of how each method generalizes under domain mismatch. Although the data generation function is known and specified, Gaussian noise is added to simulate uncertainty in the target values. This section presents the dataset generation and is followed by a comparison of model performance and an assessment of the staged B-DANN's uncertainty quantification (UQ) quality.

\subsubsection{Synthetic Dataset}
\label{subsubsec:synthetic_dataset}

To benchmark the performance of the TL strategies in a controlled manner, a synthetic dataset was generated for source and target domains. This consisted of a five-dimensional nonlinear base function, defined as follows:

\begin{align}
f(\mathbf{x}; \boldsymbol{a}, \boldsymbol{\omega})
= \kappa \Big[
a_1
+ a_2 \sin(\omega_1 x_1 + \omega_2 x_2)
+ a_3 \cos(\omega_3 x_1 x_2)
+ a_4 \log(1 + x_2^2)
+ a_5 x_3 \nonumber \\
+ a_6 x_4^2
+ a_7 x_5
+ a_8 \sin(\omega_4 x_1 x_3)
+ a_9 \cos(\omega_5 x_3 x_4)
+ a_{10} \sin(\omega_6 (x_1 + x_5)x_3)
\Big], \label{eq:src_core}
\end{align}

where \(\boldsymbol{a} = [a_1, \ldots, a_{10}]^\top\) and \(\boldsymbol{\omega} = [\omega_1, \ldots, \omega_6]^\top\) are domain-specific parameter vectors that jointly form \(\boldsymbol{\Theta} = \{\boldsymbol{a}, \boldsymbol{\omega}\}\). The scalar \(\kappa\) provides a global scaling constant. For the \textit{source domain}, inputs are used directly:

\[
y_s = f(\mathbf{x}; \boldsymbol{\Theta}_s) + \varepsilon_s, \qquad
\varepsilon_s \sim \mathcal{N}(0, \sigma_s^2).
\]

For the \textit{target domain}, a deterministic warp \(g(\mathbf{x})\) modifies the input distribution:
\[
g(\mathbf{x}) =
\begin{bmatrix}
1.2\,\sin(1.3\,x_1) + 1.5\\
x_2 + 0.4\,\cos(1.5\,x_3)\\
x_3 + 0.3\,\sin(0.8\,x_1 x_2)\\
x_4\\
x_5
\end{bmatrix},
\]
and the output is:
\[
y_t = f(g(\mathbf{x}); \boldsymbol{\Theta}_t) + \varepsilon_t, \qquad
\varepsilon_t \sim \mathcal{N}(0, \sigma_t^2).
\]

Both domains use \(\kappa = 2.0\) and noise levels \(\sigma_s = \sigma_t = 0.05\).
The parameter vectors differ slightly to induce a controlled conditional shift:
\[
\boldsymbol{\Theta}_s =
\begin{cases}
\boldsymbol{a}_s = [2.0, 5.0, 0.8, 1.0, -0.5, 0.4, -0.2, 0.6, 0.5, 0.3],\\
\boldsymbol{\omega}_s = [2.0, 1.2, 1.8, 1.5, 2.2, 1.7],
\end{cases}
\]
\[
\boldsymbol{\Theta}_t =
\begin{cases}
\boldsymbol{a}_t = [2.0, 5.0, 0.8, 1.0, -0.4, 0.35, -0.3, 0.6, 0.5, 0.3],\\
\boldsymbol{\omega}_t = [2.2, 1.0, 2.0, 1.8, 2.0, 1.9].
\end{cases}
\]

The source and target functions, when sampling from $\mathbf{X}\in [1.0,3.0]$, can produce an output value approximately in the range of $[-11,26]$, which can introduce misleading relative error metrics at values close to zero. To mitigate these concerns, a smooth quantile-based scaling was applied to all output values, mapping them to a range of $[1,5]$ using a soft sigmoid transformation. To avoid clipping at extrema, the transformation was bounded by the 5th and 95th percentiles of the combined source-target distributions; points outside of these quantiles were linearly scaled. This method of scaling avoids under representing potentially-useful outlier information, which could occur when using entirely linear mappings such as min-max normalization.

For the source set, 7,000 samples were generated using a fixed random seed and were partitioned into 5,000 train, 1,000 validation, and 1,000 test samples. The validation and test partitions are used for base model diagnostics to ensure that satisfactory performance has been reached on the source task. 

In the target domain, 1,000 samples were generated. From this target dataset, a 250-point validation partition and 250-point testing partition were set aside (totaling 50\% of the target dataset). As a training data ablation study, the training set was artificially reduced to four sizes: 75, 150, 250, and 500 points. These sizes were chosen based on prior experience with direct TL performance under data scarcity~\cite{furlong2024transfer}. The lower bound of 75 points was conservatively set to avoid total model collapse while still providing a challenging low-data scenario.

Data preprocessing was then performed for both source and target datasets. In the from-scratch models, the target training data were used to fit scalers for the standardization (z-score normalization) of all target training, validation, and testing data. In the TL-enabled models, the source and target training data were used jointly to fit the scalers. Post-standardization, each training datasets' points were shuffled to reduce the potential for biasing the model due to input position.

\subsubsection{Model Performance Metrics}
\label{subsubsec:synthetic_model_evaluation}

For each of these training dataset sizes, hyperparameter optimization was completed for the models in accordance with procedures outlined in Section \ref{subsec:hp_optimization}. Since the staged B-DANN has adversarial components, and training ablation groups continually decrease in size, training stability was a necessary consideration. To establish that these models' performance metrics are reproducible and not outliers, a simple random seed ensemble was conducted for each training strategy in each ablation group. This step simply trained each model configuration $n=20$ times, incrementing the random seed for each in a prescribed seed set. 

Following the training of each ensemble model, evaluation was performed using the common 250-point test dataset to ensure a consistent basis for comparison. For Bayesian variants, inference was performed using 200 Monte Carlo samples per input; this value was found to be sufficient for stability with a simple convergence study. The mean and variance of the 200 samples were taken for each test input to represent the point estimate and associated uncertainty for the Bayesian variants. 

For each training strategy in each ablation group, the 20 performance metric values were aggregated and recorded for their means and standard deviations. Since the goal is to directly compare the nominal performance of each configuration group, these standard deviations were used to construct 95\% confidence intervals (CIs) about each error metrics mean.

A set of six error metrics was used as a basis for comparison: the mean absolute relative error ($\upmu_{\mathrm{error}}$), the maximum absolute relative error ($\mathrm{Max}_{\mathrm{error}}$), the standard deviation of absolute relative error values ($\mathrm{Std}_{\mathrm{error}}$), the relative root mean square error (\textit{rRMSE}), the percentage of relative error values exceeding 10\% ($P_{\epsilon > 10\%}$) and the coefficient of determination ($R^2$). This collection of error metrics is intended to provide a well-rounded assessment of central accuracy, worst-case performance, and distribution of error.

The error metrics, constructed from the random seed ensemble's test means and 95\% CI values, are reported in Table \ref{tab:tl_dann_benchmark_all} and are organized by training dataset size and training strategy. Across all training sizes and evaluation metrics, the staged B-DANN approach consistently outperforms both of the other training strategies. As the number of training samples increases, all three methods increase in accuracy, as expected. Across these cases, the staged B-DANN maintains a substantial performance advantage in key central and extreme metrics. The staged B-DANN also yields performance benefits in reducing tail behavior of the error distribution, with the maximum error significantly below that of the other methods in all three cases (e.g., 38.44\% compared to approximately 52\% of the from-scratch and direct transfer methods). Similar improvements are observed in the \textit{rRMSE}, standard deviation of error, $P_{\epsilon > 10\%}$, and $R^2$ values, indicating a higher stability and lower variance in predictions. 

\begin{table}[htb!]
    \centering
    \caption{TL method test results on the synthetic conditional shift benchmark using 75, 150, 250, and 500 target training points with 95\% CIs based on 20 independent training runs.}
    \label{tab:tl_dann_benchmark_all}
    \renewcommand{\arraystretch}{1.2}
    \resizebox{\textwidth}{!}{\begin{tabular}{l|ccc|ccc|ccc|ccc}
        \toprule
        \multirow{2}{*}{Metric}
        & \multicolumn{3}{c|}{75 Points} 
        & \multicolumn{3}{c|}{150 Points} 
        & \multicolumn{3}{c|}{250 Points} 
        & \multicolumn{3}{c}{500 Points} \\
        & Scratch & Direct & Staged & Scratch & Direct & Staged & Scratch & Direct & Staged & Scratch & Direct & Staged \\
        \midrule
        $\upmu_\text{error}$ (\%)          
        & $9.56\pm0.47$ & $9.19\pm0.41$  & \cellcolor{gray!25} $6.71\pm0.21$  
        & $7.29\pm0.39$ & $6.63\pm0.38$  & \cellcolor{gray!25} $4.43\pm0.10$  
        & $5.82\pm0.33$ & $5.25\pm0.33$  & \cellcolor{gray!25} $3.61\pm0.09$  
        & $4.01\pm0.11$ & $3.54\pm0.11$  & \cellcolor{gray!25} $2.75\pm0.05$ \\

        $\text{Max}_{\text{error}}$ (\%)   
        & $52.55\pm4.15$ & $51.46\pm3.27$ & \cellcolor{gray!25} $38.44\pm2.08$
        & $38.13\pm3.68$ & $39.41\pm3.89$ & \cellcolor{gray!25} $25.93\pm2.13$ 
        & $37.57\pm4.50$ & $32.78\pm4.45$ & \cellcolor{gray!25} $19.96\pm0.77$ 
        & $20.50\pm2.28$ & $19.05\pm2.06$ & \cellcolor{gray!25} $19.00\pm1.39$ \\

        $\text{Std}_{\text{error}}$ (\%)   
        & $9.39\pm0.58$ & $9.04\pm0.52$ & \cellcolor{gray!25} $6.70\pm0.33$  
        & $7.14\pm0.52$ & $6.59\pm0.48$ & \cellcolor{gray!25} $3.94\pm0.16$  
        & $5.82\pm0.53$ & $5.12\pm0.50$ & \cellcolor{gray!25} $3.35\pm0.09$  
        & $3.50\pm0.17$ & $3.12\pm0.17$ & \cellcolor{gray!25} $2.55\pm0.08$ \\

        $rRMSE$ (\%)                       
        & $13.41\pm0.721$ & $12.89\pm0.64$ & \cellcolor{gray!25} $9.49\pm0.37$ 
        & $10.21\pm0.63$ & $9.35\pm0.59$ & \cellcolor{gray!25} $5.93\pm0.17$ 
        & $8.24\pm0.60$ & $7.34\pm0.59$ & \cellcolor{gray!25} $4.92\pm0.12$ 
        & $5.33\pm0.19$ & $4.72\pm0.19$ & \cellcolor{gray!25} $3.75\pm0.07$ \\

        $P_{\epsilon > 10\%}$ (\%)       
        & $34.08\pm2.01$ & $32.80\pm2.04$  & \cellcolor{gray!25} $21.34\pm1.07$  
        & $24.56\pm1.79$ & $21.26\pm1.84$  & \cellcolor{gray!25} $8.82\pm0.77$  
        & $17.10\pm1.70$ & $13.44\pm1.80$  & \cellcolor{gray!25} $5.42\pm0.44$  
        & $6.34\pm0.82$ & $4.36\pm0.62$  & \cellcolor{gray!25} $2.14\pm0.29$ \\

        $R^2$ (--)       
        & $0.69\pm0.03$ & $0.71\pm0.05$  & \cellcolor{gray!25} $0.85\pm0.01$  
        & $0.81\pm0.05$ & $0.84\pm0.05$  & \cellcolor{gray!25} $0.93\pm0.00$  
        & $0.88\pm0.03$ & $0.90\pm0.04$  & \cellcolor{gray!25} $0.95\pm0.00$  
        & $0.94\pm0.01$ & $0.96\pm0.01$  & \cellcolor{gray!25} $0.97\pm0.00$ \\
        \bottomrule
    \end{tabular}}
\end{table}

\begin{figure}[htb!]
    \centering
    \begin{subfigure}{0.49\textwidth}
        \centering
        \includegraphics[width=\linewidth]{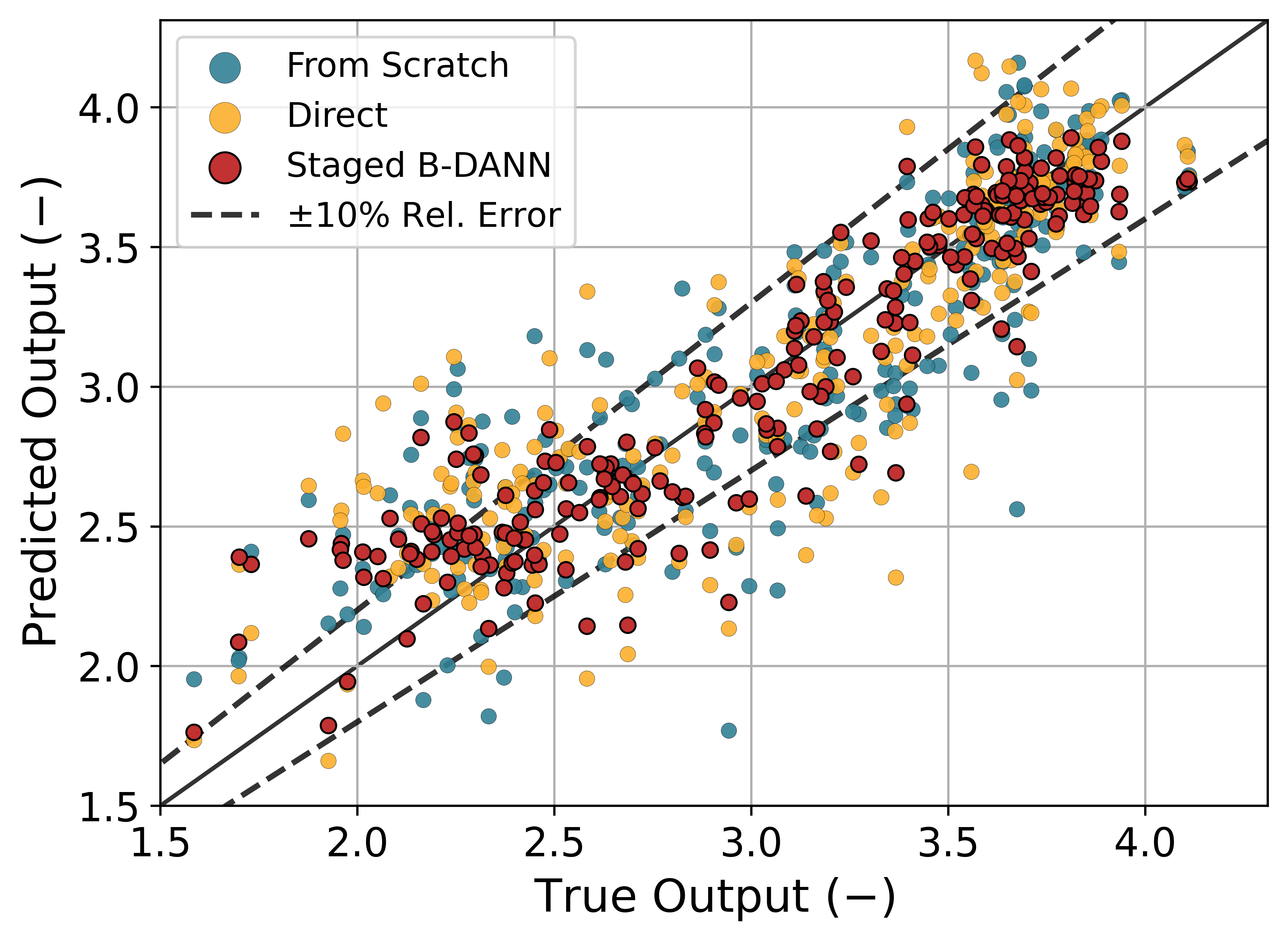}
        \caption{75 Points}
        \label{subfig:synthetic_benchmark_parity_150}
    \end{subfigure}
    \begin{subfigure}{0.49\textwidth}
        \centering
        \includegraphics[width=\linewidth]{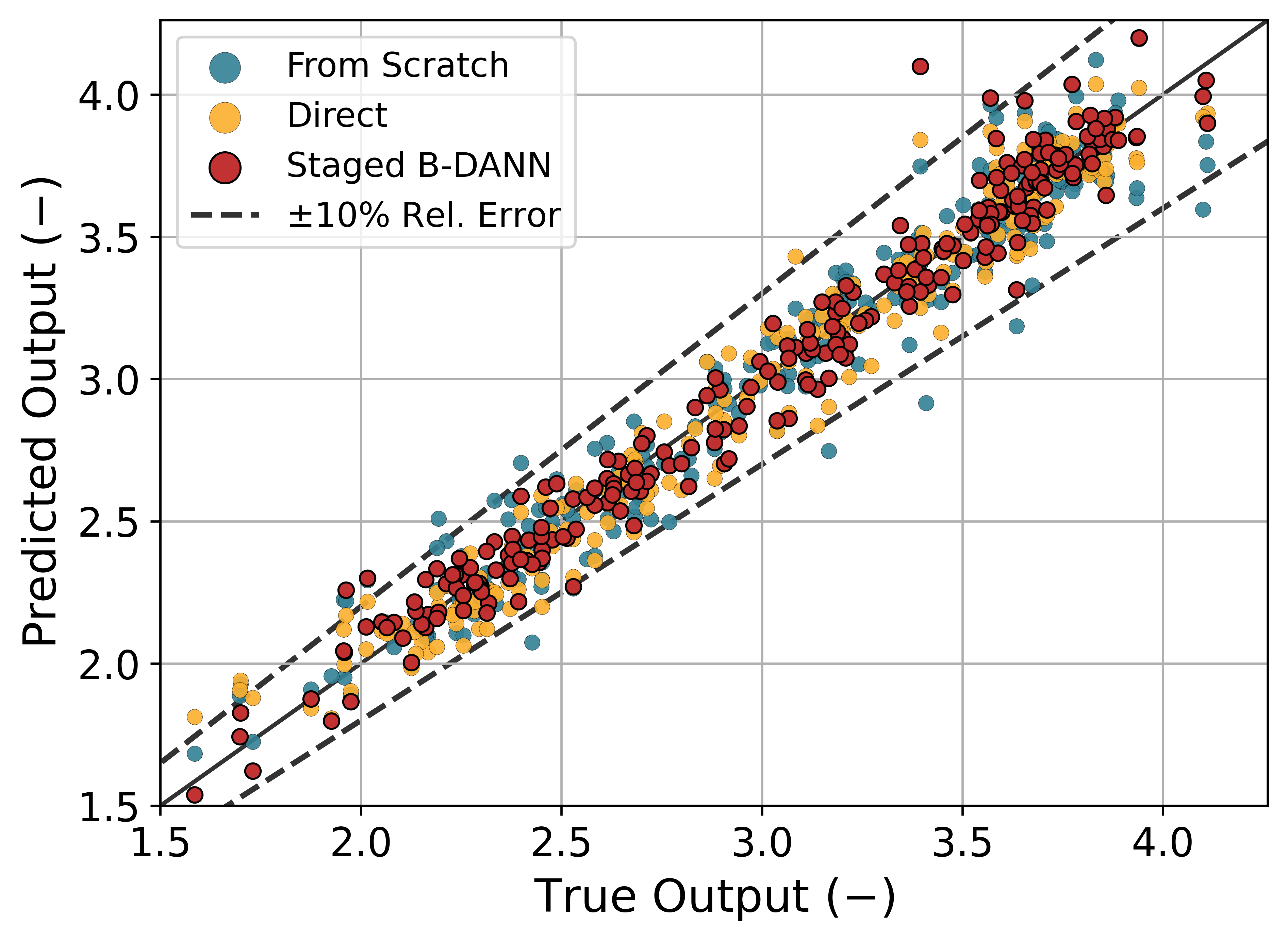}
        \caption{500 Points}
        \label{subfig:synthetic_benchmark_comparison_kde}
    \end{subfigure}
    \caption{Comparison of prediction parity in the 75-point and 500-point training data groups.}
    \label{fig:synthetic_benchmark_parity}
\end{figure}

To visualize the distribution of predictions between training strategies, parity plots are presented for the minimum and maximum ablation groups in Figure \ref{fig:synthetic_benchmark_parity}. The 75-point group sees a large spread of points, with varying levels of cohesion over the range of true outputs. With a substantially restricted training dataset size and a complex problem, it is not unexpected to see such performance degradation from data-driven models. Despite this, the staged B-DANN predictions are visibly more concentrated around the identity line compared to the other two methods. There remains the existence of high-error B-DANN predictions, but to a significantly lesser degree (21.34\% of staged B-DANN error values are beyond 10\% errors, compared to 34.08\% and 32.80\% of from-scratch and direct transfer values). The 500-point training group experienced significantly reduced spread along the identity line in all models. Despite this, the staged B-DANN predictions remain the most coherent with the exception of a high-value outlier (19.00\% error) near a true output of 3.4.

\begin{figure}[htb!]
    \centering
    \includegraphics[width=0.5\linewidth]{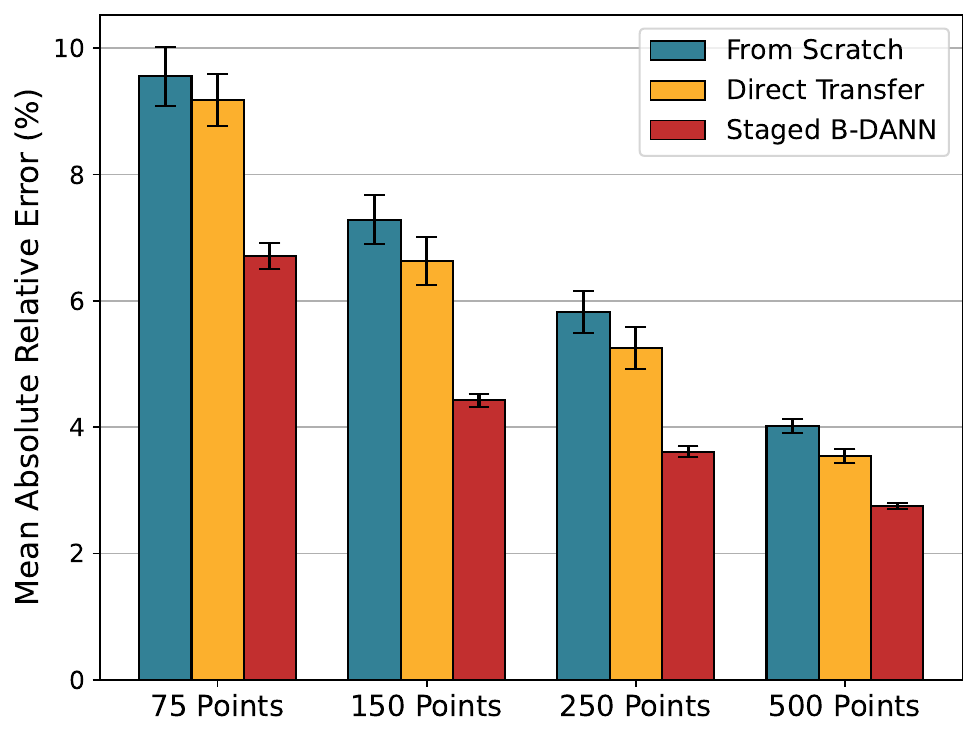}
    \caption{Direct comparison of each ML-based method's $\mu_{\text{error}}$ values with $\pm$95\% CIs, both from the aggregated 20-seed ensemble.}
    \label{fig:synthetic_benchmark_mape}
\end{figure}

To further illustrate performance differences between the three training strategies, the mean absolute relative error values are presented as bars in Figure \ref{fig:synthetic_benchmark_mape} for each of the training data sizes. As consistent with Table \ref{tab:tl_dann_benchmark_all}, the from-scratch approach yields the highest error values compared with those of the TL-based approaches. As expected, all three strategies see a rise in error with a reduction in training data. The staged B-DANN method maintains a clear performance margin at every dataset size, as well as consistently reporting the tightest 95\% CIs.

\subsubsection{Staged B-DANN Uncertainty Estimates}
\label{subsubsec:synthetic_uncertainty_estimates}

\begin{figure}[htb!]
    \centering
    \begin{subfigure}{0.495\textwidth}
        \centering
        \includegraphics[width=\linewidth]{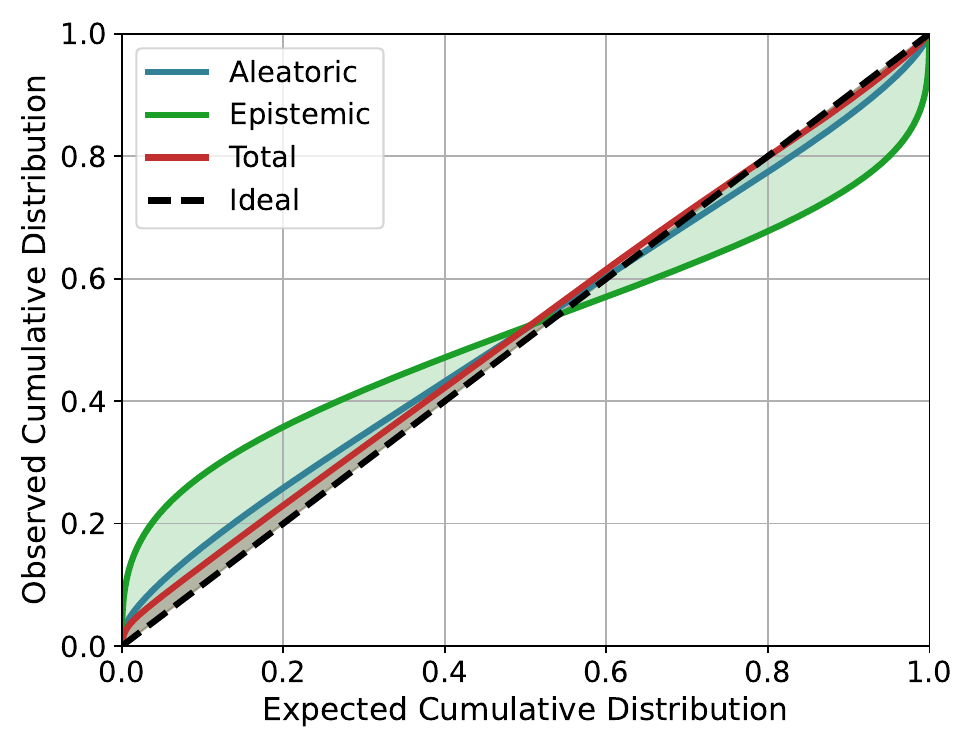}
        \caption{Uncertainty calibrations}
        \label{subfig:synthetic_benchmark_calibration_500}
    \end{subfigure}
    \begin{subfigure}{0.485\textwidth}
        \centering
        \includegraphics[width=\linewidth]{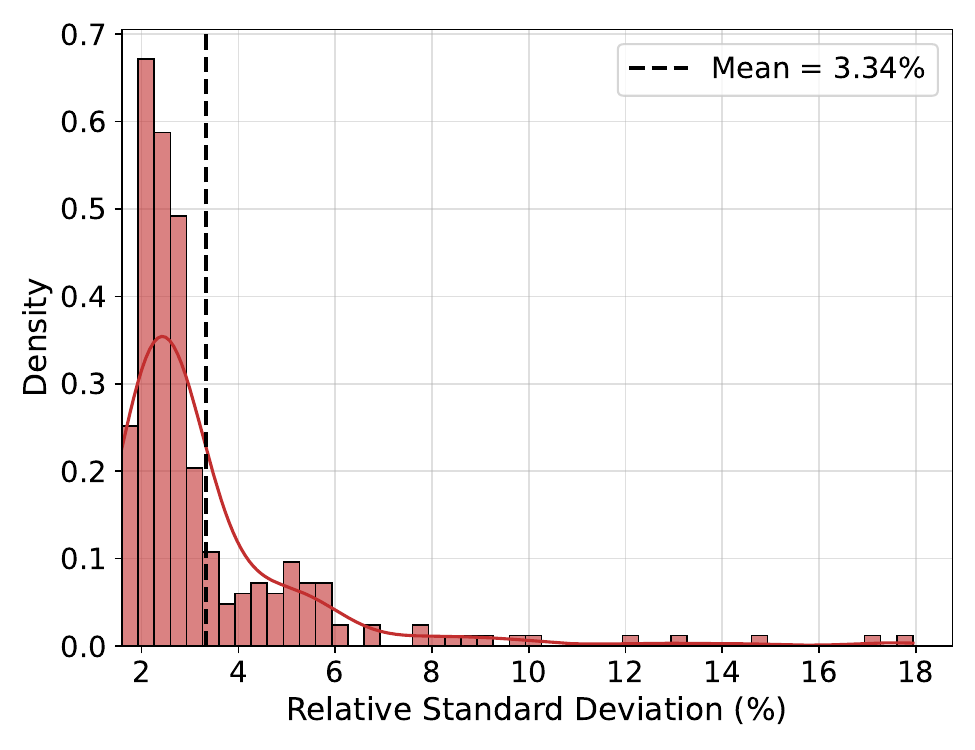}
        \caption{Total uncertainty distribution}
        \label{subfig:synthetic_benchmark_std_distribution}
    \end{subfigure}
    \caption{Analysis of uncertainty estimates produced by the 500-point staged B-DANN approach via uncertainty calibration and distribution.}
    \label{fig:synthetic_benchmark_calibration_std}
\end{figure}

One of the primary benefits of the staged B-DANN is its ability to provide principled uncertainty estimates as a result of using a BNN architecture in Stage 3. These pointwise uncertainties are particularly useful for implementing fallback mechanisms, where a reliable low-fidelity model can selectively override the ML prediction in cases of exceedingly high epistemic uncertainty. For UQ to be actionable, especially in sensitive workflows, predicted uncertainties must be both well-calibrated and accurate. Calibration curves can be constructed to compare the empirical cumulative distribution of uncertainty-normalized residuals with the ``theoretical'' standard normal distribution. This allows for the detection of systematic underprediction or overprediction of uncertainty estimates~\cite{tran2020methods}. A quantitative \textit{miscalibration area} score can also be computed, defined as the area between the calibration curve and the ideal calibration line. Figure \ref{subfig:synthetic_benchmark_calibration_500} presents the calibration curves for the estimated aleatoric, epistemic, and total uncertainties provided by the 500-point staged B-DANN model. On the left side of the plot, the epistemic curve is above the ideal line, indicating that the predicted standard deviations, which normalize the residuals, are too small. This indicates that the model is, on average, overconfident. With a miscalibration area of 0.106, this is still considered to be well-calibrated~\cite{tran2020methods}. The aleatoric uncertainty estimates are also slightly overconfident, but with a smaller magnitude and miscalibration area of 0.032. It is important to note that since variances combine nonlinearly without guarantee of a similar correlation to residuals, the total uncertainty's calibration may not be an obvious combination of the two sources' calibrations. The total uncertainty is observed to be the most well-calibrated with a slight tendency for overconfidence, with a miscalibration area of 0.018.

Assessing the distribution of predicted uncertainties is also an important step of evaluating the quality of uncertainty estimates. Distributional analysis of relative standard deviation (rStd) values across the input domain helps determine whether the model accounts for input-dependent variability. In a well-performing model, uncertainty estimates should differ in magnitudes across the input domain. Models producing uniform uncertainty estimates across all inputs are ineffective and unusable for decision-making. Figure \ref{subfig:synthetic_benchmark_std_distribution} presents a histogram and associated KDE of the relative standard deviations produced by the 500-point staged B-DANN. The distribution is right-skewed, with a tail extending to a maximum relative standard deviation value of 17.94\% and most predictions concentrated around lower values. The overall shape of these predictions suggests that the model distinguishes between different regions of the testing dataset with the expected distributional form.

The results in Figure \ref{subfig:synthetic_benchmark_calibration_500} and Figure \ref{subfig:synthetic_benchmark_std_distribution} indicate favorable uncertainty estimation behavior, supporting the reliability of the staged B-DANN's predictive distributions. Although uncertainty characteristics can vary between models, datasets, and problems, the observed calibration and distributional trends indicate that the approach produces viable and meaningful estimates in a benchmark setting. The overall performance over the course of this synthetic benchmark indicates that the staged B-DANN can quantitatively outperform both from-scratch and direct transfer training strategies in various training dataset size scenarios while maintaining high-quality uncertainty estimates. Although this performance is favorable, the synthetic nature of this benchmark motivates the need for testing in real-world problems to evaluate practical applicability.

\subsection{Case Study 2: CHF in Rectangular Channels}
\label{subsec:rectangular_channels}

Critical heat flux (CHF) is a safety-related limiting quantity in boiling systems. It marks the threshold at which a small increase in heat flux can lead to drastic and unstable increases in material temperature~\cite{todreas2021nuclear}. This directly leads to inefficiencies in the boiling process and increases stress on equipment, leading to wear and potential failure. In nuclear reactors, this quantity is especially important when considering accident conditions, such as depressurization of the reactor vessel during a coolant line rupture. Rapidly and nonuniformly increasing the local temperature on fuel assembly structures can render fuel bundles unusable or lead to a release of fission products into the primary loop. Due to these potential consequences, significant work has been performed to accurately predict CHF in various thermal hydraulic states.

The primary methods of predicting CHF rely on empirical correlations and lookup tables derived directly from experimental data. These experiments have largely been conducted in boiling tubes and have limited applicability in other geometries. Several compilations of these tube experiments are available, the largest being the U.S. Nuclear Regulatory Commission (NRC) database used to generate the 2006 Groeneveld lookup table, consisting of nearly 25,000 measurements~\cite{groeneveld2019critical}. This large amount of available tube data has allowed for extensive investigation of data-driven ML prediction methods using architectures such as DNNs, BNNs, and support vector machines~\cite{jiang2013combination}\cite{kim2021prediction}\cite{zubair2022critical}\cite{helmryd2024investigation}\cite{KHALID2024125441}\cite{khalid2024enhancing}\cite{soleimani2024analyzing}\cite{YANG2024123167}\cite{alsafadi2025predicting}\cite{ahmed2025optimized}\cite{qi2025machine}. Datasets for other experiment types (e.g., for rectangular channels), however, are sparse and lack the same level of curation, resulting in significantly less ML-focused research in those domains~\cite{kim2021prediction}\cite{NIU2024125042}\cite{furlong2024transfer}.

Rectangular geometries are frequently applied to the design of research reactors, which often use plate-type fuel~\cite{durand_1997}. This fuel type is typically constructed of thin plates of uranium-containing material and is clad in a conductive material such as aluminum. These plates are then arranged to create thin rectangular channels for coolant to flow against the heated surfaces. Most of the time, these assemblies of plates are boxed, with metal walls preventing cross-flow between interior coolant channels. Low-power reactors rarely have the capability to reach CHF, even in accident conditions, and are not typically modeled. Because these reactors are not designed to produce electricity, they do not boil their coolant during normal operation. Higher-power research reactors, such as the High Flux Isotope Reactor~\cite{cole1960high} at Oak Ridge National Laboratory and the Japan Research Reactor series~\cite{sudo1985core}, require safety analyses for various design-basis accidents in which CHF can be exceeded. To accomplish this, several correlations with various degrees of accuracy have been developed specifically for rectangular geometries.

The work performed in this study is intended to improve the accuracy of CHF predictions in rectangular channels by building both purely data-driven and hybrid residual-correction ML models while also leveraging the capabilities of the previously developed TL strategies.

\subsubsection{Hybrid CHF Modeling}

Hybrid residual-correcting ML methods have been increasingly employed for the prediction of CHF, improving interpretability, performance in data-scarce domains, and alignment to physical expectations when compared with purely data-driven methods. Introduced by Zhao et al.~\cite{zhao2020prediction}, the hybrid approach leverages domain knowledge by incorporating an established thermal hydraulic correlation into the learning process. Several studies have investigated hybrid modeling for the prediction of CHF in tube geometries with success~\cite{jin2021unified}\cite{khalid2023comparison}\cite{furlong2024hybrid}. Quantification of model uncertainties has also been studied, experimenting with various underlying ML architectures such as deep ensembles, BNNs, and deep Gaussian processes~\cite{mao2024uncertainty}\cite{furlong2025physics}. There currently is no preexisting work implementing hybrid modeling combined with TL in CHF modeling, and there is no preexisting work applying hybrid modeling to rectangular channel CHF prediction.

\begin{figure}[htb!]
    \centering
    \includegraphics[width=0.9\linewidth]{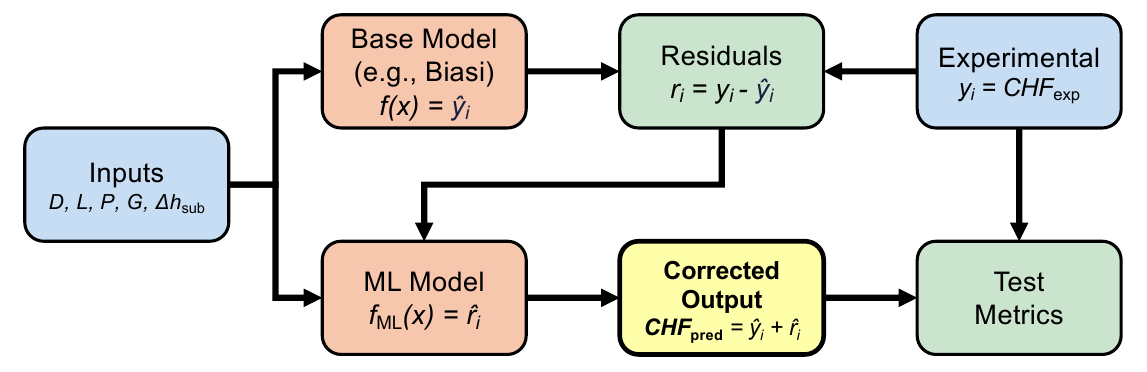}
    \caption{Overview of the hybrid model workflow in \textit{training} configuration.}
    \label{fig:hybrid_workflow}
\end{figure}

In this study, hybrid models are developed with a fixed workflow, as shown in Figure \ref{fig:hybrid_workflow}. The inputs for each CHF model are the hydraulic diameter ($D$), heated length ($L$), pressure~($P$), mass flux ($G$), and inlet subcooling ($\Delta h_{\mathrm{sub,in}}$). These inputs are first passed into a base model, which in this case is an empirical correlation providing a low-fidelity estimate ($\hat{y}_i$) of the CHF value. During training, where experimental ground truth ($y_i$) is available, a residual is computed as $r_i = y_i - \hat{y}_i$. An ML model is then trained to predict this residual $\hat{r}_i$ using the same input vector as provided to the base model. In the inference phase, where true CHF values are unknown, the corrected CHF prediction is given by $\hat{y}_i + \hat{r}_i$. This approach allows the ML model to improve the low-fidelity estimate, increasing accuracy while preserving much of the physical interpretability present in the base correlation.

\subsubsection{DeBortoli Dataset}

\begin{figure}[htb!]
    \centering
    \includegraphics[width=0.5\linewidth]{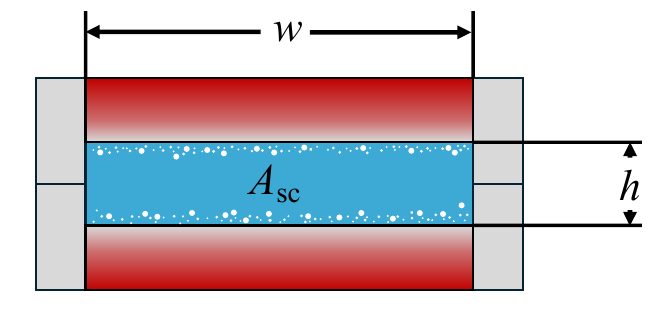}
    \caption{Generic plan view of the rectangular design used in the dataset's experiments.}
    \label{fig:rectangular_plan_view}
\end{figure}

The dataset used for the training and testing of the rectangular models originates from a series of experiments performed by R.~A.~DeBortoli et al.~\cite{debortoli1958forced} at the Bettis Thermal and Hydraulics Laboratory. Located in West Mifflin, Pennsylvania, and operated by the Westinghouse Electric Corporation, this high-pressure testing facility was composed of two operational loops at the time of the technical report: the Bettis Burnout Loop and the High-Pressure Pressure Drop Loop. The test series conducted included datasets for both tubular and rectangular geometries in various materials, flow directions, and other operational conditions. The rectangular channels were fabricated by welding the thin edges of two machined halves together, with transition segments welded on the ends. A cross-section view is give in Figure \ref{fig:rectangular_plan_view}, showing the general design and placement of heated surfaces. For a more detailed description of the testing facility, see technical report WAPD-188~\cite{debortoli1958forced}.

Each channel is geometrically defined by its height (\textit{h}), width (\textit{w}), and heated length $L_{\text{h}}$, with two opposing electrically heated planes. The vertical walls of the channel are not electrically heated and transmit less than 5\% of the total applied power. The extracted DeBortoli dataset contained experiments from four different test articles, each with unique dimensions. In all cases, the width was held at 25.40 \si{\milli\meter} (1 in.) but varied in height by a small amount. With an average \textit{h} of 2.22 \si{\milli\meter}, the heated/unheated perimeter ratio is large, on average 11.44. The hydraulic diameter ($D_{\mathrm{h}}$) was selected as the characteristic length, computed with the expression $2wh/(w+h)$.

The test procedure of the experiments was typical for CHF experiments and was carried out by continually raising the power in the channel's heating elements until a nonlinear rise in outside wall temperature was recorded, indicating burnout. The actual heat flux value was then computed by dividing the applied electrical power by the subchannel area and then multiplying by a 0.95 reduction factor. This procedure was performed on a large range of operational conditions over the course of the 400 collected test runs. The previously referenced NRC public database for tubes was selected as the source dataset due to the abundance of high-quality experimental measurements. A comprehensive list of parameter ranges for these datasets is reported in Table \ref{tab:rectangular_dataset_info}.

\begin{table}[htb!]
\centering
    \begin{threeparttable}
        \caption{Ranges of experimental conditions for the NRC tube database and the DeBortoli~\cite{debortoli1958forced} rectangular channel dataset. The outlet equilibrium quality is denoted by $x_{\mathrm{e,cr}}$, and $q''_{\mathrm{cr}}$ represents the CHF. Values for $G$ and $q''_{\mathrm{cr}}$ are provided with the precision reported in the original text. Note that the upper value in a given parameter cell indicates the \textbf{maximum}, and the lower value indicates the \textbf{minimum}.}
        \begin{tabular}{lccccccccc}
        \toprule
          Dataset & $D_{\mathrm{h}}$ & $L$ & $P$ & $G$ & $\Delta h_{\mathrm{sub,in}}$ & $x_{\mathrm{e,cr}}$ & $q''_{\mathrm{cr}}$ \\
           & (\si{\milli\meter}) & (\si{\meter}) & (\si{\mega\pascal}) & (\si{\kilo\gram\per\square\meter\per\second}) & (\si{\kilo\joule\per\kilo\gram}) & (-) & (\si{\kilo\watt\per\square\meter})\\ \midrule
          NRC~\cite{groeneveld2019critical} & 16.00 & 20.00 & 20.00 & 7{,}964 & 1{,}644  &  0.99 & 16{,}339 \\
          (Source) & 2.00 & 0.05 & 0.01   & 8   & $-1{,}211$   & $-0.50$ & 50  \\ \midrule
          DeBortoli~\cite{debortoli1958forced} & 4.66 & 0.69 & 13.79  & 6{,}487 & 1{,}521.50\tnote{a}  &  0.99 & 7{,}772 \\
          (Target) & 2.42 & 0.15 & 4.12   & 23   & 28.41\tnote{a}    &  0.01 & 151  \\
        \bottomrule
        \end{tabular}
        \begin{tablenotes}
        \item[a] Values derived from other parameters and IAPWS-IF97 fluid properties.
        \end{tablenotes}
        \label{tab:rectangular_dataset_info}
    \end{threeparttable}
\end{table}

\begin{figure}[htb!]
    \centering
    \includegraphics[width=0.6\linewidth]{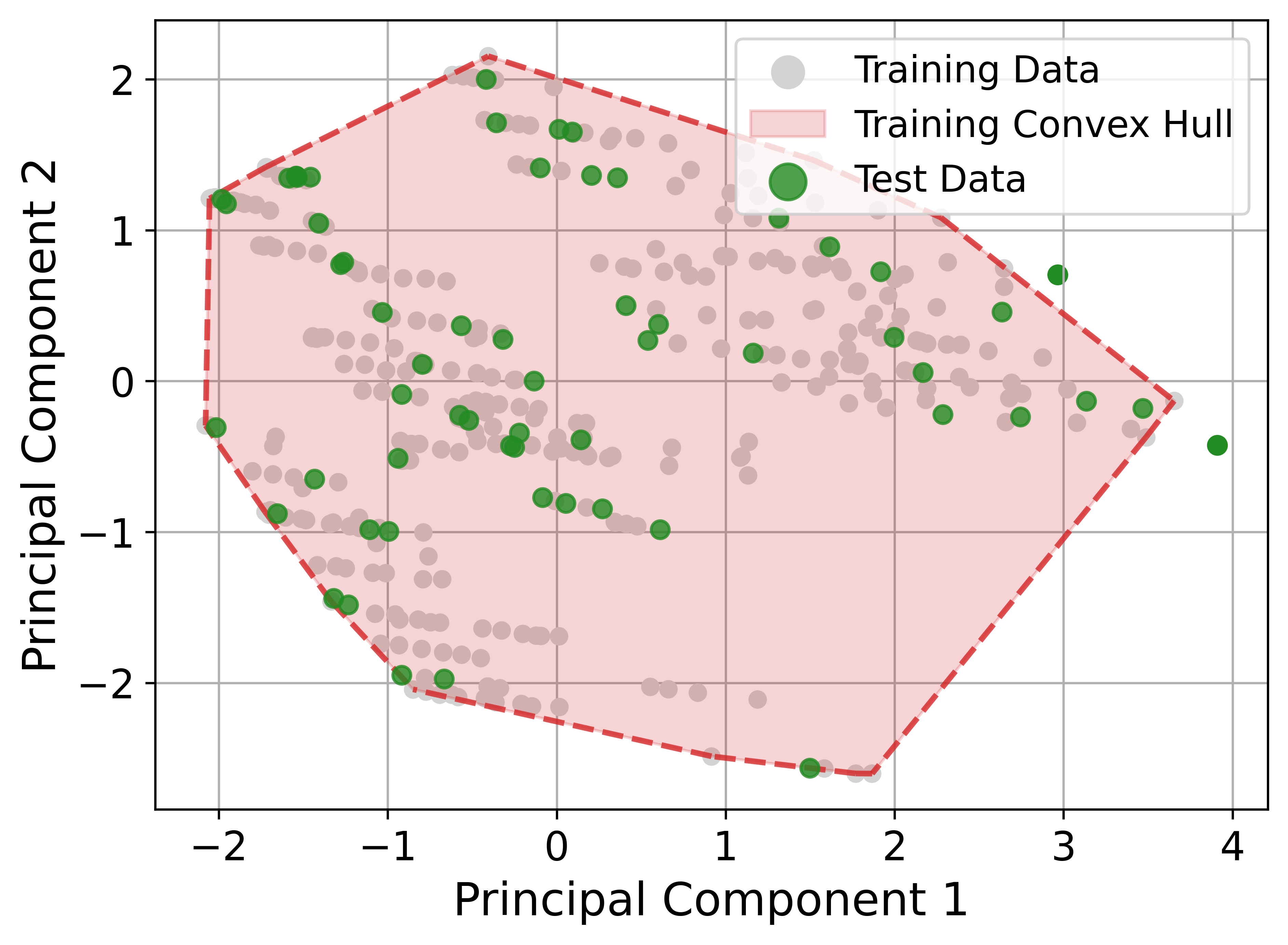}
    \caption{Two-component PCA visualization with convex hull analysis depicting the testing dataset compared with the training dataset. The test points are well-distributed across the training domain, except for a region of sparsity affecting both test and training points.}
    \label{fig:rectangular_pca}
\end{figure}

Data preprocessing was conducted using the procedure outlined in Section \ref{subsec:synthetic_benchmark}, but an 80\%/5\%/15\% split was used to increase the available test points. Due to the limited number of data available for this study, care was taken to ensure that the test cases were representative of the training domain. To visualize the distribution of evaluation points relative to training data, a 2D principal component analysis (PCA) was performed, as shown in Figure \ref{fig:rectangular_pca}. Although PCA involves dimensionality reduction and some information loss, it is a valuable tool to inspect the coverage and structure of the dataset in lower-dimensional space. 

To assess whether the trained model would be subject to test points outside the training domain (extrapolation), a convex hull was constructed. A convex hull is the smallest convex set that fully encloses the training points. In two dimensions, it can be visualized as the tightest rubber band stretched around the outermost training points. In Figure \ref{fig:rectangular_pca}, the green test points are distributed across the training domain without apparent bias toward any region. There is a region of sparsity identified in the lower right region of the convex hull, with two points outside of the hull indicating an extrapolation region. Although a two-dimension PCA is desired for visual interpretation, a full-dimensional convex hull analysis is required to definitively identify all extrapolating points. In this analysis, 17 of the 60 (28.3\%) test points were found to be outside the training domain (in the extrapolatory regime).

\subsubsection{CHF Correlations}
\label{subsubsec:chf_correlations}

A set of empirical correlations was compared against a purely data-driven ML model and the correlations' hybrid ML equivalents. Both the Biasi and Bowring correlations are used due to their widespread usage in various thermal hydraulic codes. These correlations were built using data from round tube experiments but are still used as-is when modeling rectangular channels in these codes, such as in CTF~\cite{salko2020ctf}. The rectangular channel dataset used in this study also included two new correlations for rectangular channels. Because the goal of the below study is to evaluate and enhance the \textit{generalization} of CHF models, these two new correlations are not used, as doing so could lead to misleading results if overfit to the calibration dataset. To provide a comparison against a rectangular-channel-specific correlation built on rectangular channel data, the Sudo--Kaminaga (S-K) correlation was chosen~\cite{sudo1985experimental}. Of the correlations present in rectangular channel literature, the S-K model is frequently present and demonstrates strong performance~\cite{bertocchi2024predicting} upon evaluation.

With the three base correlations chosen (Biasi, Bowring, and S-K), the test partition of the dataset was used to evaluate the efficacy of their predictions. This partition is the same for all models tested in this study to provide a fair comparison. The results of each correlation's performance are included in Table \ref{tab:rectangular_correlation_metrics} using five metrics. Both of the round-tube-based correlations show mean absolute relative error values below 20\%, whereas the S-K correlation's predictions have a value of 47.76\%. This result is unexpected because the model derived using the rectangular channel was expected to outperform both Biasi and Bowring on rectangular channel data. This performance difference is continued through the other four metrics, with 95\% of S-K's predictions falling outside a 10\% relative error envelope. Both the Biasi and Bowring also suffer in the same metric, with values of 73\% and 78\%, respectively.

\begin{table}[htb!]
    \centering
    \caption{Comparison of selected empirical correlations evaluated on the rectangular database's testing subset.}
    \label{tab:rectangular_correlation_metrics}
    \begin{tabular}{lccc}
        \toprule
        Metric & Base & Base & Base \\
         & Biasi & Bowring & S-K \\
        \midrule
        $\upmu_\text{error}$ (\%) & \cellcolor{gray!25}$15.59$ & $19.97$ & $47.76$ \\ \midrule
        $\text{Max}_{\text{error}}$ (\%) & \cellcolor{gray!25}$33.73$ & $45.60$ & $86.07$ \\ \midrule
        $\text{Std}_{\text{error}}$ (\%) & \cellcolor{gray!25}$8.96$ & $11.01$ & $21.17$ \\ \midrule
        $rRMSE$ (\%) & \cellcolor{gray!25}$17.98$ & $22.80$ & $52.24$ \\ \midrule
        $P_{\epsilon > 10\%}$ (\%) & \cellcolor{gray!25}$73.33$ & $78.33$ & $95.00$ \\
        \bottomrule
    \end{tabular}
\end{table}

\begin{figure}[htb!]
    \centering
    \includegraphics[width=0.6\linewidth]{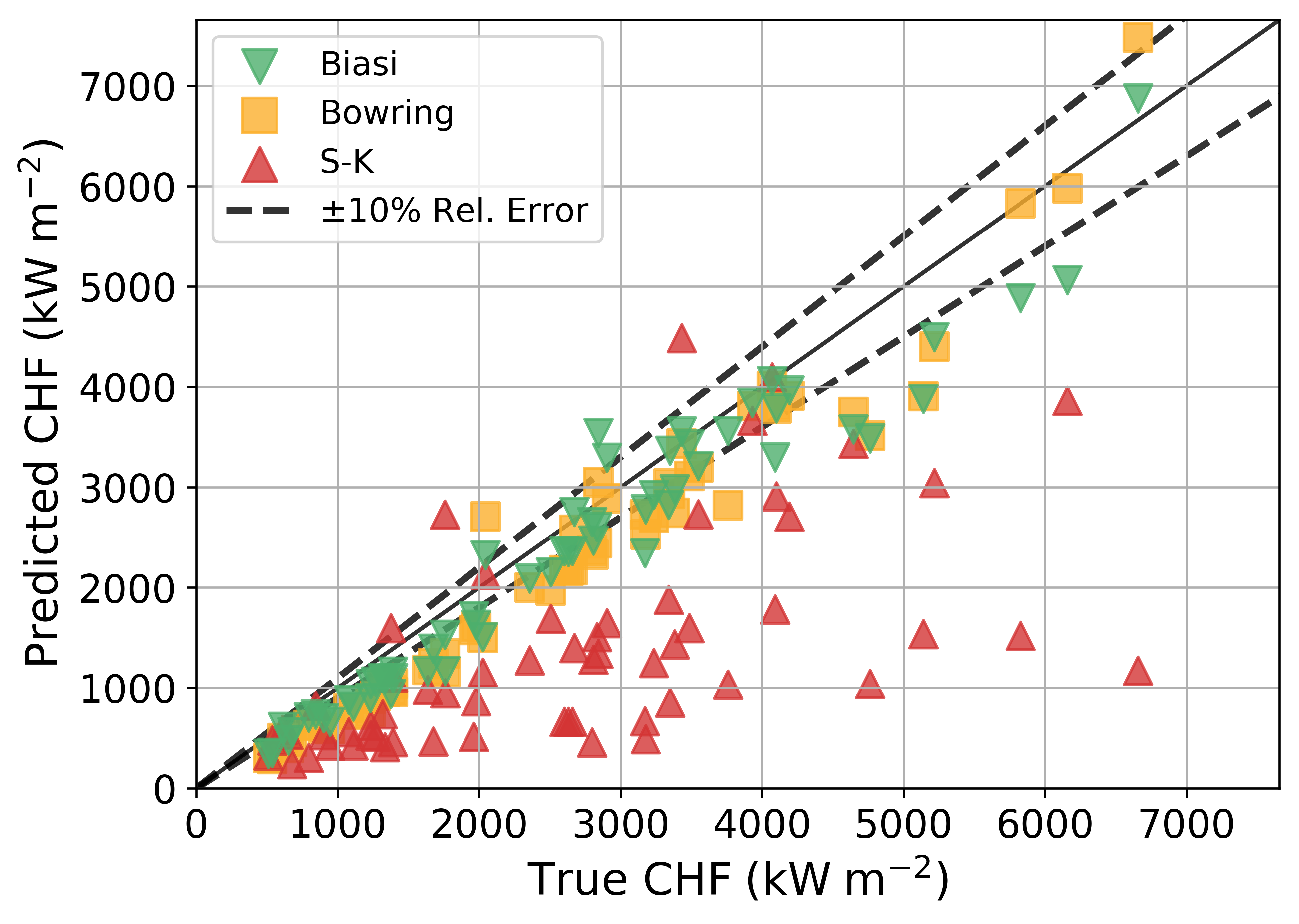}
    \caption{Parity of the standalone correlations' predictions compared to the 60-point testing subset of the rectangular database.}
    \label{fig:rectangular_correlation_parity}
\end{figure}

To gain a better view of the error distribution, a parity plot was constructed (Figure \ref{fig:rectangular_correlation_parity}) to view the individual predictions from each model in relation to the corresponding true values. A cursory inspection immediately shows the S-K correlation's large deviation from true parity, consistently underpredicting CHF, in one case even by 6,000 \si{\kilo\watt\per\square\meter}. The distribution possesses a large variance and spread over the course of the predictions, indicating a generally poor fit of the model in this dataset's domain. It should be noted that there is more favorable behavior at smaller CHF values but still with sizable deviation from the identity line.

Both the Biasi and Bowring correlations produce predictions more similar to experimental values compared than the S-K predictions, with a smaller variance and a more coherent pattern. A pattern of CHF underprediction is still evident; nearly all points are below the identity line, and most are outside the 10\% error envelope. Considering both the quantitative data from Table \ref{tab:rectangular_correlation_metrics} and the qualitative distribution in Figure \ref{fig:rectangular_correlation_parity}, none of the three correlations are considered to be effective at predicting CHF values for the rectangular channels of this dataset.

\subsubsection{ML Model Evaluation}

The four ML-based models, one purely data-driven model and three hybrid models, were optimized and trained using the same procedure defined in Section \ref{subsec:synthetic_benchmark}. To assess run-to-run training stability, 20 models were initially trained per configuration with varied random seeds to construct 95\% CIs about the error metric mean values. Instability was noted particularly in the case of the from scratch and direct transfer models, with unexpectedly little in the case of the staged B-DANN models. To increase the confidence in finding nominal error metric values (effectively population means), 100 models were then trained for each configuration. The resulting mean performance metrics and corresponding 95\% CIs are reported in Table \ref{tab:chf_rectangular_table_master}. The first characteristic of note is the global reduction in mean absolute relative error values and those above a value of 10\% compared to the standalone correlations. The staged B-DANN in particular improves performance from the best standalone correlation (Biasi) in every metric in all prediction methods except for the maximum relative error and the pure ML $\mathrm{Std}_{\mathrm{error}}$. All training methods in all prediction method groups are observed with deficiencies in the maximum relative error, all of which are attributed to a single point at a true CHF value of 151 \si{\kilo\watt\per\square\meter}. From a run-to-run perspective, this ``difficult'' point had the greatest variability, for example ranging from an error value of 33\% to 267\% over the 100 from scratch pure ML runs. Since this point's error value can affect outlier-sensitive metrics, $P_{\epsilon > 10\%}$ values become important; every ML-based model exhibits fewer points outside 10\% absolute error compared to the correlations.

\begin{table}[htb!]
    \centering
    \caption{Comparison of training strategies and prediction methods for the CHF case study.}
    \label{tab:chf_rectangular_table_master}
    \renewcommand{\arraystretch}{1.2}
    \resizebox{\textwidth}{!}{\begin{tabular}{l|ccc|ccc|ccc|ccc}
        \toprule
        \multirow{2}{*}{Metric}
        & \multicolumn{3}{c|}{Pure ML} 
        & \multicolumn{3}{c|}{Hybrid Biasi} 
        & \multicolumn{3}{c|}{Hybrid Bowring} 
        & \multicolumn{3}{c}{Hybrid S-K} \\
        & Scratch & Direct & Staged & Scratch & Direct & Staged & Scratch & Direct & Staged & Scratch & Direct & Staged \\
        \midrule
        $\upmu_\text{error}$ (\%)          
        & $11.45\pm0.45$ & $9.27\pm0.42$ & \cellcolor{gray!25} $7.43\pm0.20$
        & $7.45\pm0.14$ & $6.87\pm0.16$ & \cellcolor{gray!25} $5.03\pm0.07$
        & $8.14\pm0.16$ & $7.80\pm0.19$ & \cellcolor{gray!25} $5.64\pm0.10$
        & $11.09\pm0.40$ & $8.96\pm0.34$ & \cellcolor{gray!25} $6.28\pm0.14$ \\

        $\text{Max}_{\text{error}}$ (\%)   
        & $123.02\pm11.78$ & $100.02\pm9.58$ & \cellcolor{gray!25} $77.78\pm8.00$
        & $79.21\pm4.58$ & $71.43\pm5.48$ & \cellcolor{gray!25} $47.32\pm1.72$
        & $96.54\pm5.27$ & $88.60\pm6.73$ & \cellcolor{gray!25} $57.87\pm3.41$
        & $111.52\pm9.99$ & $82.85\pm8.33$ & \cellcolor{gray!25} $54.09\pm4.02$ \\

        $\text{Std}_{\text{error}}$ (\%)   
        & $18.06\pm1.40$ & $14.54\pm1.13$ & \cellcolor{gray!25} $11.01\pm0.89$
        & $11.13\pm0.51$ & $10.13\pm0.61$ & \cellcolor{gray!25} $6.82\pm0.19$
        & $13.10\pm0.61$ & $12.34\pm0.77$ & \cellcolor{gray!25} $8.16\pm0.37$
        & $16.97\pm1.17$ & $12.62\pm0.99$ & \cellcolor{gray!25} $8.06\pm0.42$ \\

        $rRMSE$ (\%)                       
        & $21.52\pm1.39$ & $17.35\pm1.15$ & \cellcolor{gray!25} $13.40\pm0.84$
        & $13.44\pm0.48$ & $12.30\pm0.61$ & \cellcolor{gray!25} $8.49\pm0.18$
        & $15.47\pm0.58$ & $14.68\pm0.73$ & \cellcolor{gray!25} $9.95\pm0.35$
        & $20.39\pm1.17$ & $15.58\pm0.98$ & \cellcolor{gray!25} $10.26\pm0.40$ \\

        $P_{\epsilon > 10\%}$ (\%)       
        & $33.88\pm1.39$ & $26.57\pm1.21$ & \cellcolor{gray!25} $22.85\pm0.79$
        & $19.57\pm0.79$ & $18.33\pm0.77$ & \cellcolor{gray!25} $12.68\pm0.48$
        & $22.23\pm0.85$ & $21.85\pm0.81$ & \cellcolor{gray!25} $13.83\pm0.59$
        & $32.92\pm1.23$ & $27.95\pm1.30$ & \cellcolor{gray!25} $18.25\pm0.70$ \\

        $R^2$ (--)       
        & $0.96\pm0.00$ & $0.97\pm0.00$ & \cellcolor{gray!25} $0.98\pm0.00$
        & $0.97\pm0.00$ & $0.97\pm0.00$ & \cellcolor{gray!25} $0.99\pm0.00$
        & $0.98\pm0.00$ & $0.98\pm0.00$ & \cellcolor{gray!25} $0.99\pm0.00$
        & $0.96\pm0.00$ & $0.97\pm0.00$ & \cellcolor{gray!25} $0.98\pm0.00$ \\
        \bottomrule
    \end{tabular}}
\end{table}

Comparing the training methods within each of the prediction method groups, the staged B-DANN predictions consistently outperform both the from-scratch and direct transfer training strategies in all metrics. The magnitudes of these performance differences are relatively similar across groups, with the largest observed in the hybrid S-K models (10.26\% \textit{rRMSE} compared to 15.58\% of the direct transfer configuration). The direct transfer can be observed with clear but smaller performance gains compared to the from scratch models; a limited number metrics overlap in 95\% CIs, particularly in the hybrid Biasi and hybrid Bowring groups. The largest difference between the direct transfer and from scratch approaches appears in the hybrid S-K group, which also sees heightened error metrics compared to the other two hybrid approach groups. A potential cause of this increased error could simply be due to the standalone S-K correlation's relatively poor performance compared to the Biasi and Bowring correlations, providing less relevant data regarding the target (rectangular) domain.

\begin{figure}[htb!]
    \centering
    \begin{subfigure}{0.48\textwidth}
        \centering
        \includegraphics[width=\linewidth]{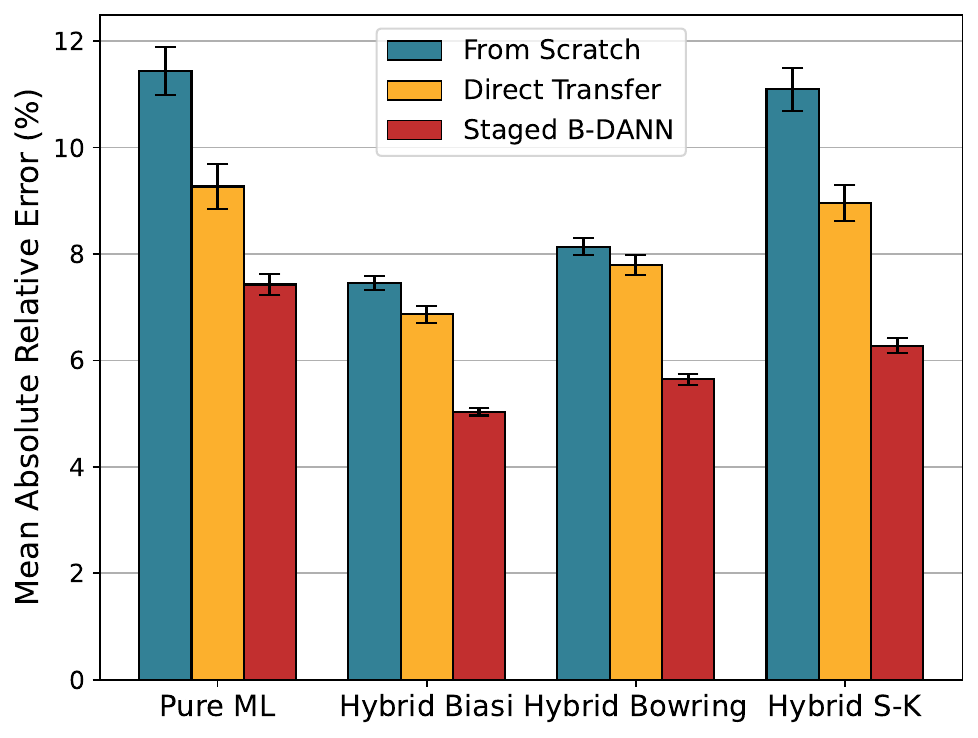}
        \caption{$\mu_{\mathrm{error}}$}
        \label{subfig:rectangular_channel_bar_mape}
    \end{subfigure}
    \begin{subfigure}{0.48\textwidth}
        \centering
        \includegraphics[width=\linewidth]{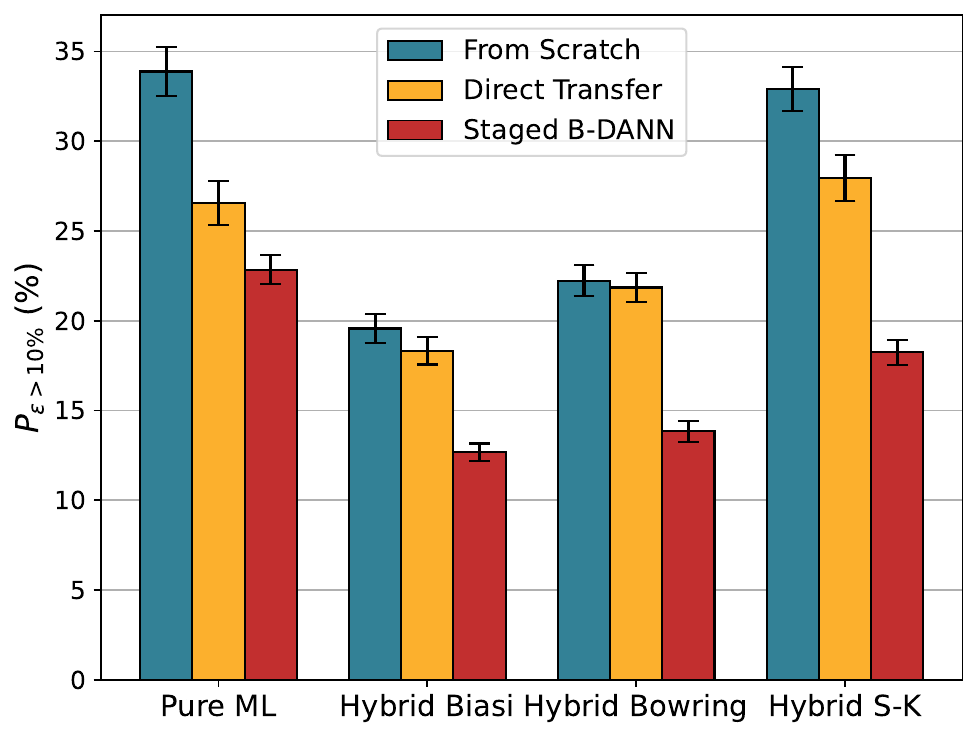}
        \caption{$P_{\epsilon>10\%}$}
        \label{subfig:rectangular_channel_bar_pctape10}
    \end{subfigure}
    \caption{Direct comparison of each ML-based method's $\mu_{\text{error}}$ and $P_{\epsilon>10\%}$ values, illustrating differences in error distribution \textbf{central tendency} and \textbf{spread}.}
    \label{fig:rectangular_channel_bar_comparison}
\end{figure}

As a visualization of performance between methods, bar charts of mean absolute relative error values and the percentage of absolute relative error values above 10\% are provided in Figure \ref{fig:rectangular_channel_bar_comparison}. The observed trends reaffirm that in this problem, the staged B-DANN training strategy significantly improved the central tendencies and spreads of error distributions in all cases. The variability between prediction method groups is evident, with the largest staged B-DANN error easily identified in the pure ML training method group. This visual comparison also demonstrates the greater run-to-run instability of the from scratch and direct transfer strategy between methods, as observed by the significantly larger 95\% CIs. Out of all prediction methods, the hybrid Biasi performs the most favorably in error and stability, reflected in both mean values and 95\% CIs.

\section{Conclusions}
\label{sec:conclusions}

This work introduced and validated a three-stage TL approach that integrates parameter transfer and shared latent space transfer. This approach begins with pretraining a DNN on the source domain to initialize a feature extractor with representations suited to a domain neighboring the target. Domain alignment is then performed using a modified DANN, which allows the feature extractor to find domain-invariant representations by attempting to confuse a domain label classifier. These aligned features are used to initialized a BNN, which is then fine-tuned on the target domain to improve performance under conditional shift and provide uncertainty estimates.

The staged B-DANN method was first benchmarked with a synthetic dataset, which was generated for both source and target domains with conditional shift. The performance of the trained-from-scratch, direct transfer (parameter transfer), and staged B-DANN models was compared in three different target training dataset sizes. The staged B-DANN routinely outperformed the other methods. Additional supporting evaluation was provided by parity visualizations. Calibration diagrams and predictive distribution analyses further indicated that the Stage 3 BNN provided reliable and well-behaved uncertainty estimates. These findings show that the staged B-DANN framework offers benefits in both accuracy and UQ, even under domain shift and data scarcity.

A study was then conducted considering the prediction of CHF in rectangular channels using both purely data-driven and hybrid residual-correction ML models. Three empirical correlations served as both baseline comparators and low-fidelity models for the hybrid approaches. From-scratch, direct transfer, and staged B-DANN models were developed; the TL strategies used the abundant tube-based Groeneveld NRC database as the source domain and the DeBortoli rectangular-based database as the target domain. Out of all ML-based prediction groups, the hybrid Biasi models performed the most favorably in key metrics. In terms of training methods, the staged B-DANN models improved ML-based performance in all groups with greater run-to-run stability, often with a significant margin.

Across both synthetic and CHF studies, the staged B-DANN consistently outperformed both from scratch and direct transfer training strategies, yielding superior accuracy and providing well-calibrated uncertainty estimates. Although sensitivity to hyperparameters in adversarial schemes remains an open limitation, the staged B-DANN was observed to have stable training and consistent performance over both studies. Future work will focus on further optimization of the staged B-DANN methodology and extending its application to more complex, high-dimensional physical modeling problems.

\section*{Acknowledgments}
This work was partially funded by the US Department of Energy (DOE) Office of Nuclear Energy Distinguished Early Career Program (DECP) under award number DE-NE0009467.
This work was also partially supported by the Nuclear Energy Advanced Modeling and Simulation Program for the modeling and simulation of nuclear reactors under DOE contract no.~DE-AC05-00OR22725.

\newpage
\bibliography{./bibliography.bib}

\end{document}